\documentclass[runningheads]{llncs}

\usepackage{eccv}

\usepackage{eccvabbrv}

\usepackage{graphicx}
\usepackage{booktabs}
\usepackage{enumitem}
\usepackage{tabularx}
\usepackage{multirow}

\usepackage{color}
\usepackage{soul}

\usepackage[accsupp]{axessibility}  

\usepackage{hyperref} 
\hypersetup{colorlinks=true, allcolors=black, urlcolor=eccvblue}

\usepackage{orcidlink}

\usepackage{pifont}
\usepackage{xcolor}
\usepackage{colortbl} 
\usepackage{wrapfig}

\definecolor{lt_few}{HTML}{D7FFDA}
\definecolor{lt_tail}{HTML}{FCD7FF}
\definecolor{lt_head}{HTML}{D7EBFF}
\definecolor{lt_rareact}{HTML}{FFE5CC}

\newcommand{\cmark}{\ding{51}}
\newcommand{\xmark}{\ding{55}}

\newcommand{\paragraphcustom}[1]{\vspace{4pt}\noindent\textbf{#1}}

\usepackage{multibib}
\newcites{suppl}{Supplementary References}

\usepackage{tcolorbox}
\tcbuselibrary{breakable, skins}
\usepackage{listings}

\definecolor{promptbg}{HTML}{F7F7F8}
\definecolor{promptframe}{HTML}{BBBBBB}

\newtcolorbox{promptbox}[1][]{%
  enhanced, breakable,
  colback=promptbg, colframe=promptframe,
  boxrule=0.5pt, arc=2pt,
  left=6pt, right=6pt, top=4pt, bottom=4pt,
  fonttitle=\bfseries,
  title={#1}
}

\lstdefinestyle{json}{
  basicstyle=\ttfamily\scriptsize,
  breaklines=true,
  columns=fullflexible,
  keepspaces=true,
}

\newtcolorbox{actionprofile}[1]{
  colback=promptbg, colframe=promptframe,
  boxrule=0.5pt, arc=2pt,
  fonttitle=\bfseries, title={#1},
  breakable, left=4pt, right=4pt, top=4pt, bottom=4pt
}

\newcommand{\fillthresh}{B} 

\begin{document}

\title{Gen2Balance: Generative Balancing for Long-Tailed Video Action Recognition}

\titlerunning{Gen2Balance}

\author{Prajwal Gatti\inst{1} \quad
Simon Jenni\inst{2} \quad
Fabian Caba Heilbron\inst{2} \quad
Dima Damen\inst{1}
}

\authorrunning{P.~Gatti et al.}

\institute{$^1$University of Bristol\quad \quad $^2$Adobe Research \\\vspace*{6pt}
\url{https://prajwalgatti.github.io/gen2balance/}}

\maketitle

\begin{abstract}
We address the problem of training on long-tailed data for video action recognition. We propose to augment the training set using a text-to-video generative model, conditioned on diverse text prompts grounded in action profiles and training exemplars.
Our approach, called \textit{Gen2Balance}, converts an imbalanced training set into a balanced combination of real and generated video clips. 
To effectively learn from such data, we employ a two-stage training strategy that mitigates domain shift and yields significant improvements.

We evaluate on long-tailed versions of standard benchmarks: UCF-101 (UCF-LT) and a 100-class subset of Kinetics (K100-LT) selected to prioritise temporally challenging actions.
Gen2Balance improves accuracy over the strongest baselines for long-tailed learning by 5.1\% and 7.0\% on the respective datasets.
On rare actions from the RareAct dataset (\eg, \textit{cut keyboard}), Gen2Balance improves accuracy by 31.9\%, demonstrating effectiveness for scarce actions. By varying the amount of synthetic data added, we show that partial balancing already achieves 79\% of the performance gains at 27\% of the compute cost on K100-LT, highlighting the practical scalability of Gen2Balance.

\vspace{-19pt}
\end{abstract}

\begin{figure*}[!t]
  \centering
  \includegraphics[width=\linewidth]{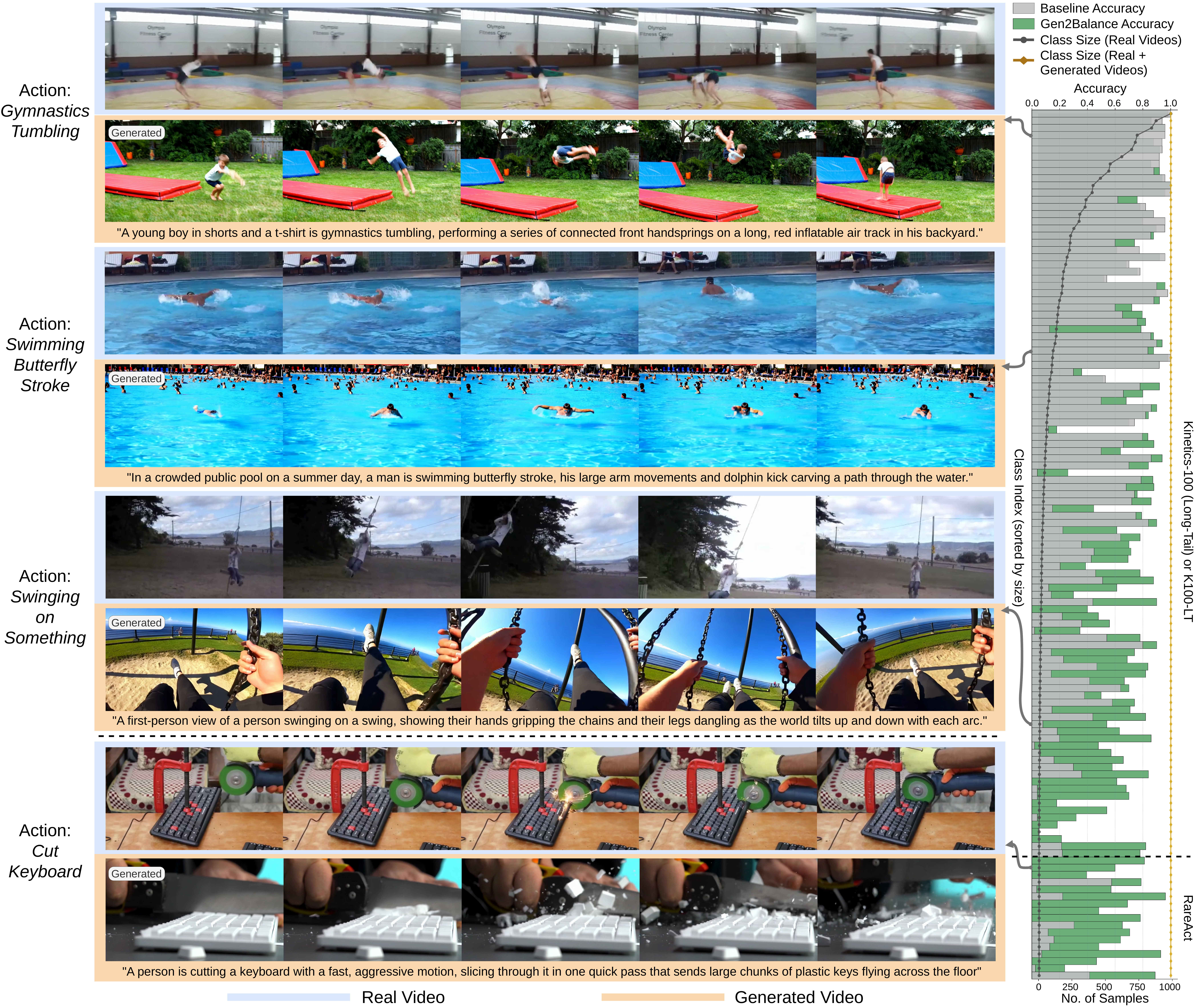}
  \caption{We introduce \textbf{Gen2Balance}, a generative balancing approach for long-tail video action recognition. \textbf{(Left)} For four actions (three from K100-LT, one from RareAct~\cite{miech2020rareact}), we show a real video from the training dataset alongside a generated sample with its conditioning prompt (bottom text), illustrating the visual diversity produced by our pipeline. \textbf{(Right)} Class-wise accuracy on K100-LT and RareAct classes, sorted by training-set size (black line). The baseline degrades sharply towards the tail and few-shot classes; Gen2Balance, trained with both real and generated videos, improves few-shot class accuracy by +38.8\% (+48.0\% on RareAct classes).}
  \vspace*{-12pt}
  \label{fig:intro}
\end{figure*}

\section{Introduction}
\label{sec:intro}

Despite the increase in available data, the underlying class distribution often remains skewed and heavily long-tailed, rendering training on imbalanced data a persistent, fundamental challenge.
Existing approaches aim to improve optimisation or augment the imbalanced training data.
However, the performance on long-tailed benchmarks still lags considerably behind that of balanced or nearly balanced datasets.
For example, state-of-the-art accuracy reaches only 80.6\% on ImageNet-LT~\cite{zhao2024ltgc} and 58.0\% on CIFAR-100-LT~\cite{yu2022coca}, compared to 91.0\%~\cite{du2023global} and 89.3\%~\cite{cubuk2019autoaugment} on the balanced versions of these respective datasets, using the same backbone.
A similar gap persists in video action recognition: on UCF-101~\cite{soomro2012ucf101}, we show accuracy drops from 92\% to 64\% under long-tailed imbalance. 

The emergence of powerful generative models has naturally prompted researchers to question:
\textit{can we synthesise sufficiently diverse examples of under-represented classes to balance a long-tailed distribution?}
In the image domain, the answer has been increasingly affirmative, with recent work training on fully synthetic clones of standard datasets like ImageNet~\cite{sariyildiz2023fake,he2023is} or augmenting the training data with generated images~\cite{shin2023fill,zhao2024ltgc}.
In this work, we address long-tailed \textit{video} action recognition using generative models: a direction that, unlike in the image domain, remains largely underexplored.

This setting fundamentally differs from the image domain because video generation must also capture temporal dynamics across frames, and current models are frequently criticised for producing implausible physics, inconsistent object permanence, and unnatural motion~\cite{bansal2025videophy2,gu2025phyworldbench,thozhiyoor2025objects,motamed2025generative}. Beyond these quality concerns, generated videos must also faithfully depict the intended action class. Despite these challenges, we find that for recognition, the current generative models---when carefully prompted---produce videos of sufficient quality to achieve state-of-the-art performance in long-tailed video action recognition.

\noindent Our contributions are as follows:
\begin{itemize}[topsep=0pt, itemsep=0pt, partopsep=0pt, parsep=0pt, leftmargin=*]
    \item[$\bullet$] We propose Gen2Balance, a framework to address long-tailed video action recognition by balancing datasets with synthetic videos from a pre-trained text-to-video generative model (\cref{fig:intro}).
    \item[$\bullet$] We introduce an automatic pipeline that synthesises videos conditioned on diverse, class-faithful prompts generated by a multimodal LLM. Using this pipeline with WAN~2.1 and Gemini~2.5~Pro, we generate and publicly release a complementary dataset of 140K synthetic videos with detailed text prompts spanning 223 action classes across UCF-LT, K100-LT, and RareAct~\cite{miech2020rareact}.
    \item[$\bullet$] We compare training strategies on combined real and augmented samples. We employ a two-stage training (combined then real) and show that training with a class-balanced loss, using margins derived from real-data frequencies rather than augmented frequencies, achieves the best results.
    \item[$\bullet$] On long-tailed splits Kinetics (K100-LT) and UCF-101 (UCF-LT), we demonstrate that Gen2Balance surpasses the strongest baselines for learning with long-tailed data by up to +6.7\% on K100-LT and +5.5\% on UCF-LT few-shot classes, with gains of +31.9\% on rare actions from RareAct.
\end{itemize}

\section{Related Work}
\label{sec:related-work}

\paragraphcustom{Long-Tailed Learning}. 
Long-tailed learning approaches generally fall into three categories. \textbf{Re-sampling} methods address class imbalance by over-sampling the tail~\cite{chawla2002smote,gupta2019lvis} or under-sampling the head~\cite{liu2008exploratory}, risking overfitting or discarding valuable head-class data, respectively. \textbf{Re-weighting} strategies penalise tail-class errors more heavily by weighting the loss by class-size~\cite{ren2020balanced,cui2019classbalancedloss,tan2020equalization,tan2021equalization}, sample difficulty~\cite{deng2021pml,jamal2020rethinking,li2022equalized,lin2017focal,shu2019meta}, or logit adjustment~\cite{menon2021logit,tao2023local,li2022long,tian2020posterior,wu2021adversarial,xu2023livt}, though amplifying tail-class gradients commonly degrades head-class performance. Other strategies include decoupling representation from classifier training~\cite{kangdecoupling, zhou2020bbn,li2020overcoming,sun2025rethinking,alshammari2022long}, label smoothing~\cite{sun2025rethinking,zhong2021improving}, and multi-expert ensembles~\cite{cai2021ace,wang2020long,cui2022reslt,li2021trustworthy,zhang2022self,hou2025a}. \textbf{Data augmentation} methods address the fundamental lack of tail-class samples, either \textit{implicitly} or \textit{explicitly}. \textit{Implicit} augmentation approaches include feature space interpolation~\cite{zhang2018mixup,chou2020remix,verma2019manifold}, transferring head-variance to tail prototypes~\cite{perrett2023use,li2024feature}, hand-crafted transformations~\cite{ahn2023cuda,wang2024kill,wang2024llmautoda} or mixing in the input space~\cite{yun2019cutmix}; however, these cannot produce new information (\eg, novel scenes or object appearances). \textit{Explicit} augmentation via web-retrieval~\cite{long2022retrieval, sidhu2025search,iscen2023improving,zhao2025learning} injects new knowledge, but is prone to label noise and limited by tail-class data availability online. Our work instead leverages generative models to synthesise \textit{on demand}, targeting missing diversity without relying on web availability.

\paragraphcustom{Long-Tailed Video Recognition}. While long-tailed learning is well studied for images~\cite{liu2019large,cui2019classbalancedloss,kangdecoupling,menon2021logit,lin2017focal}, fewer works have addressed long-tailed video recognition, which introduces additional challenges of temporal reasoning. 
\cite{zhang2021videolt} introduced VideoLT, a long-tailed recognition benchmark, and proposed to dynamically resample frames based on average precision during training. 
\cite{perrett2023use} proposed new properties for characterising long-tailed data and introduced Long-Tail Mixed Reconstruction (LMR), an implicit augmentation strategy that linearly combines head and tail class features.
MOVE~\cite{moon2023move} also augments features through dynamic extrapolation within instances and frequency-based interpolation.
MEDC~\cite{hu2023medc} adopts a multi-expert approach where separate
branches model the long-tailed, uniform, and reversed distributions,
transferring head-class knowledge to tail classes. MEID~\cite{li2023meid} extends this to handle frame-level imbalance.
\cite{hu2025majority} synthesises tail-class samples in the feature space, conditioned on attention-weighted head-class features.
All these prior video works operate \textit{at the feature or logit level}, relying on pre-extracted features or older encoders.
In contrast, Gen2Balance synthesises novel videos in the pixel-space using a text-to-video generative model, injecting new appearance and motion diversity rather than combining features from the training set.

\paragraphcustom{Generative Data Augmentation}. Advances in generative models for images~\cite{rombach2022high,raisinghani2025nanobanana,batifol2025flux,midjourney2022} and videos~\cite{openai2025sora2,google2025veo3,runway2025gen45,wan2025wan} have opened a new data augmentation paradigm for recognition. 
In balanced image classification, training on fully synthetic~\cite{sariyildiz2023fake} or augmenting with synthetic data~\cite{azizi2023synthetic,he2023is} yields competitive performance, even improving robustness to biases~\cite{singh2024synthetic}.
In long-tailed settings, SYNAuG~\cite{ye2025synaug} generates synthetic images to balance the class distribution and uses MixUp to close the synthetic-to-real domain gap. Fill-Up~\cite{shin2023fill} instead personalises generation via textual inversion~\cite{gal2022image}, but inversion requires per-class training \textit{before} any generation---a prohibitive bottleneck for memory-intensive video models\footnote{We do not compare to textual inversion (\eg, Fill-Up~\cite{shin2023fill}), as training per-class inversion embeddings at the scale of WAN~2.1-14B exceeds H100 memory limits.}. Both SYNAuG and Fill-Up employ a two-stage training strategy.
LTGC~\cite{zhao2024ltgc} avoids per-class training by using an LLM to generate diverse prompts for tail-class image generation, and further uses generations to improve a vision-language model.
In videos, Li~\etal~\cite{li2025role} study generative augmentation for zero- and few-shot learning by pretraining purely on synthetic clips, then fine-tuning on real data.
In contrast, Gen2Balance combines LLM-driven prompting with training-free video generation and a classifier training strategy that learns useful representations from synthetic data while mitigating domain shift.

\vspace{-4pt}
\section{Gen2Balance Method}
\label{sec:approach}

Gen2Balance consists of two components. First, we synthesise videos to fill imbalanced classes up to a target threshold (\cref{subsec:gen2bal-generation}). To produce diverse, class-faithful videos, we use a multimodal LLM pipeline to create text prompts by using diversity axes, action profiles, and real video exemplars.
Second, we train the recognition model in two stages: learning from augmented data, followed by rehearsal on the real data (\cref{subsec:gen2bal-training}).
We first formalise the problem setup.

\subsection{Preliminaries}
We address the problem of long-tailed video action recognition. Let the training set be $\mathcal{D}_{train}{=
}\{(x_i,c_i)\}_{i=1}^{N}$, where $x_i\in\mathcal{X}$ is a video clip, $c_i \in\mathcal{C}{=}\{1,\ldots,C\}$ is the class label, and $N_j {=} |\{(x_i, c_i) \in \mathcal{D}_{train} : c_i = j\}|$ denotes the number of samples in class $j$. When classes are ordered by cardinality ($N_1 \geq N_2 \geq \ldots \geq N_C$), the imbalance ratio in the dataset~\cite{cui2019classbalancedloss} is defined as $\mathcal{I} = \frac{N_1}{N_C}$, and generally $\mathcal{I} \gg 1$ for long-tailed benchmarks. Following prior work~\cite{perrett2023use}, classes in $\mathcal{D}_{train}$ are partitioned into three groups based on sample frequency: 
\sethlcolor{lt_head}
(i)~\hl{\textit{head classes}}, which collectively account for 50\% of the training samples;
\sethlcolor{lt_few}(ii)~\hl{\textit{few-shot classes}}, each containing fewer than 20 samples; and (iii)~the remaining intermediate 
\sethlcolor{lt_tail}
\hl{\textit{tail classes}}. Along with $\mathcal{D}_{train}$, $\mathcal{D}_{test}$ is typically a class-balanced set, so that class-average accuracy equals overall accuracy.

We define the recognition model $f_{\theta}:\mathcal{X} \to \mathbb{R}^{C}$, parametrised by $\theta$. When $f_{\theta}$ is trained on $\mathcal{D}_{train}$ with standard empirical risk minimisation, without accounting for the class imbalance, the model's performance on $\mathcal{D}_{test}$ tends to be biased towards the head classes at the expense of tail and few-shot classes. 

 \begin{figure*}[!t]
  \centering
  \includegraphics[width=\linewidth]{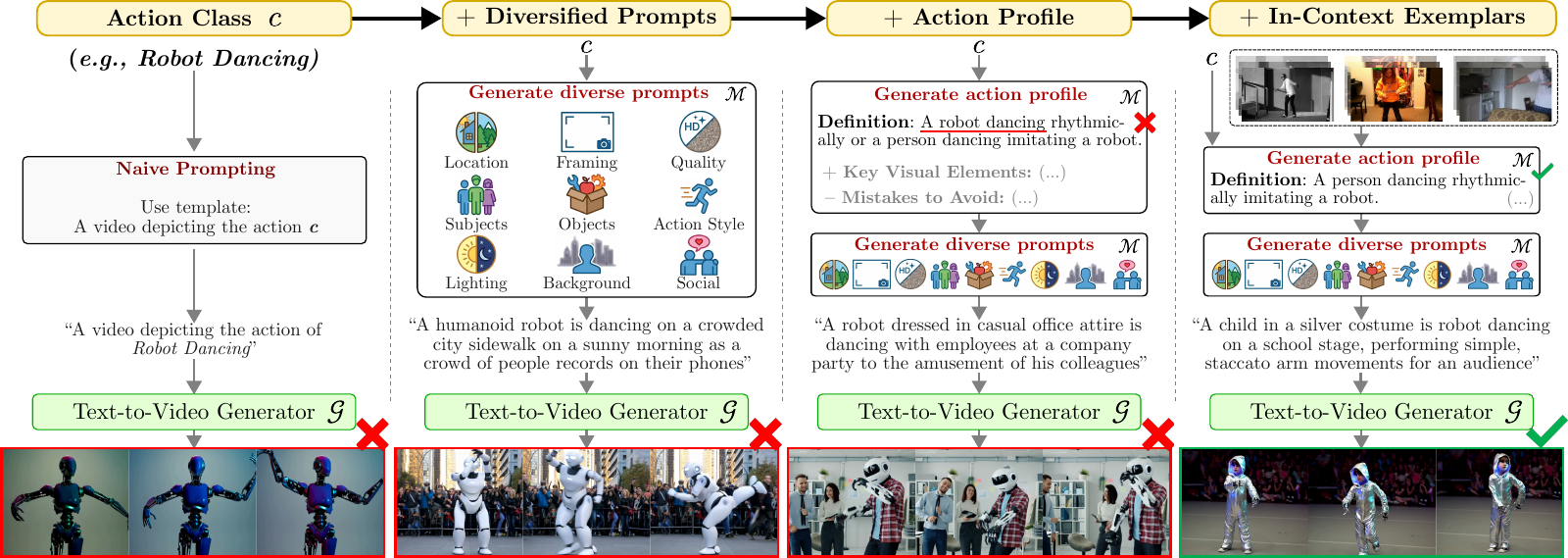}
  \caption{ We progressively illustrate our video generation pipeline on the semantically ambiguous Kinetics class \textit{Robot Dancing} (a human dance style imitating a robot). Naive prompting and diversification alone incorrectly generate mechanical robots dancing (Cols.~1--2). Adding an Action Profile (Col.~3) improves the definition, but it remains ambiguous (shows \emph{both} humans and robots dancing). By providing real exemplars to disambiguate the action profile (Col.~4), we synthesise a human dancing like a robot.}
  \vspace*{-12pt}\label{fig:method_generation_pipeline}
\end{figure*}

\subsection{Generative Filling of the Long-Tail}
\label{subsec:gen2bal-generation}
 We define a target filling threshold $\fillthresh$ such that for any class $c$ with real sample count $N_c < \fillthresh$, we generate $N'_c = \fillthresh - N_c$ synthetic videos to supplement the existing samples; classes with $N_c \geq \fillthresh$ are left untouched. In our results, we fully balance the datasets by setting $\fillthresh = N_1$, \ie, the size of the largest class. We also ablate partial filling ($\fillthresh<N_1$), which leaves classes with $N_c \geq \fillthresh$ untouched, and overfilling ($\fillthresh> N_1$), which adds synthetic data even to head classes.

 We assume access to a text-conditioned
video generative model~$\mathcal{G}$ and condition it on $\mathcal{T}_c$ text prompts where
$|\mathcal{T}_c| = N'_c$.
Next, we explain how we generate the per-class text prompts $\mathcal{T}_c$.
 
Traditional action recognition datasets~\cite{kay2017kinetics} are curated by crawling the web for video clips matching a predefined action list and filtering them based on semantic definitions of the actions. While effective, this process is constrained by the web data availability and inherits its biases---limited diversity in actor demographics, environments, and social contexts. Our objective is to construct a prompt set $\mathcal{T}_c$ for each class~$c$ that introduces controlled diversity.

A naive approach is to prompt $\mathcal{G}$ with a fixed template such as \textit{``A video depicting the action of [class name]''}, analogous to templated prompting in image-based augmentation~\cite{sariyildiz2023fake,he2023is}. This produces an inadequate training distribution for two reasons: (1) the class name alone can be semantically ambiguous or unclear, and (2) templated prompting yields repetitive, stereotypical samples that fail to capture the visual diversity needed for robust training. We address both issues through a structured pipeline, as illustrated in \cref{fig:method_generation_pipeline}.

\paragraphcustom{Diversifying Text Prompts}. Following prior work on LLM-driven text prompt generation~\cite{zhao2024ltgc,li2025role}, we leverage a multimodal large language model $\mathcal{M}$ to generate prompts that vary along nine diversity axes: (1)~environments, (2)~camera framing, (3)~video quality, (4)~actor demographics, (5)~associated props, (6)~action intensity, (7)~lighting conditions, (8)~background density, and (9)~social context. Full prompt provided in Supp~\ref{sec:supp_prompt_diversity}.

However, diversification alone does not resolve the semantic ambiguity in class names. For example, the class \textit{``Robot Dancing''} in Kinetics refers to a human dance style that mimics a robot, yet prompts produce videos of mechanical robots dancing---confusing rather than assisting the classifier (\cref{fig:method_generation_pipeline}, Col.~2).

\paragraphcustom{Disambiguating with Action Profiles}. To resolve such ambiguities, we aim to \textit{reverse-engineer the dataset curation process} by encoding the prior knowledge a human annotator would use when judging whether a video belongs to class $c$. We prompt $\mathcal{M}$ with the class name to generate an Action Profile $\mathcal{A}_c$: a textual specification comprising (a)~a \textbf{definition} of the action, (b)~\textbf{positive constraints} (key visual features that should be present), and (c)~\textbf{negative constraints} (common misconceptions to avoid). $\mathcal{M}$ then generates the diversified prompts $\mathcal{T}_c$ conditioned on both $c$ and $\mathcal{A}_c$; \ie,  $\mathcal{T}_c = \mathcal{M}(\mathcal{A}_c, c)$, steering generation towards the intended semantics. However, for semantically ambiguous classes, the profile may still adopt the wrong interpretation
(\cref{fig:method_generation_pipeline}, Col.~3).

\paragraphcustom{Grounding with In-Context Exemplars}. 
To resolve ambiguity, we supply $\mathcal{M}$ with a few-shot set $\mathcal{S}_c$ of training videos from $\mathcal{D}_{train}$ as in-context exemplars when generating the action profile $\mathcal{A}_c$.
We assume $\mathcal{M}$ is a VLM that can accept exemplar video clips.
This stage grounds the profile in the correct class semantics (\cref{fig:method_generation_pipeline}, Col.~4). Crucially, exemplars inform only the profile $\mathcal{A}_c$, not the prompts directly, \ie, $\mathcal{A}_c = \mathcal{M}(\mathcal{S}_c, c)$ and $\mathcal{T}_c = \mathcal{M}(\mathcal{A}_c, c)$. This separation prevents the limited exemplars from narrowing prompt diversity, which remains governed by the explicit axes described above.
Sample action profiles and prompts for querying $\mathcal{M}$ are provided in the Supp~\ref{supp:action-profiles} and~\ref{sec:supp_prompts}, respectively.

\paragraphcustom{Text-to-Video Generation}.
For each textual description~$\tau \in \mathcal{T}_c$, we sample a synthetic clip $\hat{x}\sim\mathcal{G}(\tau)$ and assign it label~$c$, assuming that $\tau$ is sufficiently descriptive for $\mathcal{G}$ to produce a video faithful to~$c$.

We denote the generated dataset by $\mathcal{D}_{gen} = \{(\hat{x}_i, c_i)\}$, and the augmented training set by $\mathcal{D}_{aug}=\mathcal{D}_{train}\cup\mathcal{D}_{gen}$. Our goal is to optimise $f_{\theta}$ on $\mathcal{D}_{aug}$ to learn robust representations and perform well across all classes.

\subsection{Training Gen2Balance}
\label{subsec:gen2bal-training}
\begin{figure*}[!t]
  \centering
  \includegraphics[width=\linewidth]{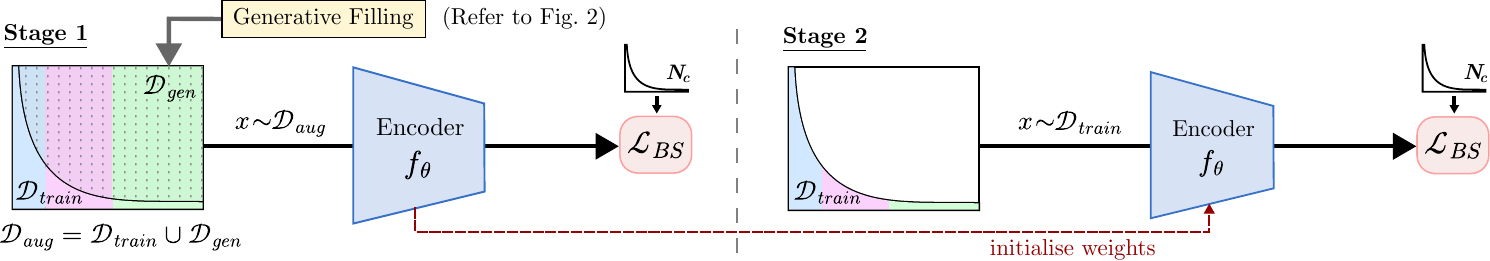}
  \caption{\textbf{The Gen2Balance Training Strategy.} Stage~1 trains on the filled dataset ($\mathcal{D}_{aug}$) with Balanced Softmax loss~($\mathcal{L}_{BS}$), using margins based on real-data frequencies ($N_c$). Stage~2 fine-tunes only on real data with the same loss to rectify domain shift.}
  \vspace*{-12.5pt}
  \label{fig:method_training}
\end{figure*}

While generative filling balances the dataset, directly training on $\mathcal{D}_{aug}$ (\ie, using standard cross-entropy without any long-tail adjustment) introduces new challenges. First, \textit{overfitting to synthetic data:} The model may learn shortcuts from synthetic samples rather than generalisable features. Second, \textit{domain shift:} Systematic differences between real and generated videos (\eg, video quality, texture artefacts, physical inconsistencies) can pull representations away from the real-data manifold. These challenges are documented in prior work~\cite{kangdecoupling,zhong2021improving,alshammari2022long}, including settings with generative augmentation~\cite{shin2023fill,li2025role,ye2025synaug}, where separating representation learning from classifier adjustment is standard practice. We employ a two-stage approach as illustrated in Fig.~\ref{fig:method_training}.

\paragraphcustom{Stage 1: Learning from Augmented Data}.
We first fine-tune $f_\theta$ on $\mathcal{D}_{aug}$ using the Balanced Softmax loss~\cite{ren2020balanced}, which adjusts loss margins based on class frequency $N_c$ to penalise errors on classes with fewer samples more heavily - $\mathcal{L}_{BS} = -\log \left( \frac{N_c e^{\eta_c}}{\sum_{j=1}^{C} N_j e^{\eta_j}} \right)$
where $\boldsymbol{\eta} = f_\theta(x) \in \mathbb{R}^{C}$ is the logit vector, $\eta_c$ its $c$-th component, and $N_c$ denotes the raw sample count of class $c$ in $\mathcal{D}_{train}$. When $\mathcal{D}_{aug}$ is fully balanced (\ie, $\fillthresh$ is set to the largest head class size), all classes would have the same count. Using total frequency counts ($N_c + N_c'$) in the loss nullifies the re-weighting effect, reducing it to standard cross-entropy. This is undesirable because the generated data, while useful, remains a noisy approximation of the true distribution (see ablations in Section~\ref{sec:experiments}). Thus, we use the real frequencies~($N_c$), which create larger loss margins for the tail and few-shot classes, so that the model uses abundant generated data for feature learning while maintaining decision boundaries calibrated to the real class priors.

\paragraphcustom{Stage 2: Rehearsal on Real Data}.
We fine-tune $f_\theta$ exclusively on $\mathcal{D}_{train}$ with the same Balanced Softmax loss. We use a reduced learning rate to make small corrective updates without catastrophic forgetting of the learnt representations.

\section{Experiments}
\label{sec:experiments}
\vspace{-5pt}

\paragraphcustom{Datasets.}
We focus our experiments on the two most widely used video action recognition benchmarks: Kinetics~\cite{kay2017kinetics} and UCF-101~\cite{soomro2012ucf101}\footnote{We exclude low-quality video datasets~\cite{goyal2017something}, which exhibit a larger synth-real gap and egocentric datasets~\cite{damen2022rescaling} as current generative models struggle with this viewpoint.}.
Following~\cite{liu2019large,perrett2023use,lin2024vlg}, we construct long-tailed versions by sampling class sizes from a Pareto distribution. We retain the size of the largest class while sampling randomly the smallest few-shot class to a minimum of 5 examples. 
For Kinetics, we select 100 classes, prioritising temporal classes, \ie, action classes where temporal information is necessary for recognition (as identified in~\cite{sevilla2021only}), to ensure the benchmark tests motion understanding.
We refer to the resulting long-tailed benchmarks as K100-LT and UCF-LT, and use the original balanced test sets for evaluation. 

We additionally evaluate on RareAct~\cite{miech2020rareact}---a dataset of actions formed by unlikely co-occurring verb-noun compositions (\eg, \textit{drill phone}, \textit{microwave shoes}). 
By design, these probe compositional generalisation~\cite{li2024c2c} over rare verb–noun compositions, here in a \emph{few-shot} regime.
Due to annotation noise, we manually curate a clean subset of 22 rare actions, ensuring no overlap between source videos in the train and test clips. 
We combine these classes with K100-LT, and append them to the few-shot partition (with 5 training samples each).
 
\begin{table}[t]
  \caption{
    \textbf{Dataset Statistics.}
    Comparison of the balanced benchmarks against our proposed long-tailed (LT) splits.
    $\mathcal{I}$: imbalance ratio ($N_{\max}/N_{\min}$);
    \textbf{H\,\%\,/\,F\,\%}: percentage of classes in the Head\,/\,Few-shot partitions ($\le20$ samples).
    Our LT splits introduce long-tail imbalance while preserving the original head-class frequencies.
    Test sets are shared across splits (balanced). $^\dagger$Proposed LT split.
  }
  \vspace*{-6pt}
  \label{tab:dataset_stats_full}
  \centering
  \scriptsize
  \setlength{\tabcolsep}{3.5pt}
  \begin{tabular}{@{}l l r r r r r >{\columncolor{lt_head}}r >{\columncolor{lt_few}}r@{}}
    \toprule
    Source & Split & \# Cls & \# Videos & $N_{\max}$ & $N_{\min}$ & $\mathcal{I}$ & H\% & F\% \\
    \midrule
    Kinetics-100 & Train & 100 & 61{,}621 & 990 & 252 & 4 & 35 &  0 \\
    K100-LT$^\dagger$ & Train & 100 & 14{,}439 & 990 &   5 & 198 & 11 & 30 \\
    Kinetics-100 & Test & 100 & 4{,}952 & \multicolumn{5}{c}{\textemdash} \\
    \midrule
    UCF-101 & Train & 101 & 9{,}537 & 121 & 72 &  2 & 45 &  0 \\
    UCF-LT$^\dagger$  & Train & 101 & 3{,}026 & 121 & 5 & 24 & 18 & 55 \\
    UCF-101 & Test & 101 & 3{,}783 & \multicolumn{5}{c}{\textemdash} \\
    \bottomrule
  \end{tabular}
  \vspace*{-12pt}
\end{table}

\paragraphcustom{Statistics and Metrics.} Table~\ref{tab:dataset_stats_full} summarises the statistics of our proposed splits. 
K100-LT has an imbalance ratio of 198 with only 11\% head classes and 30\% few-shot classes; UCF-LT has an imbalance ratio of 24 with 55\% few-shot classes. For evaluation, we report class-wise Accuracy (C/A) across three splits: head, tail, and few-shot (marked ``Few'' in tables), along with the average C/A.

\noindent \textbf{Generation ($\mathcal{M}$ and $\mathcal{G}$)}. We instantiate $\mathcal{M}$ as Gemini 2.5 Pro~\cite{comanici2025gemini} and $\mathcal{G}$ as WAN~2.1-14B~\cite{wan2025wan}. We set $|S_c|=5$ in-context exemplars for action profiles.

Datasets are fully balanced by setting $\fillthresh$ to the maximum class size (990 for K100-LT and 121 for UCF-LT).
Videos are generated as 4-second clips at 16 FPS, with 480$\times$832 resolution, using a 5.0 guidance scale and 50 inference steps.

Full balancing requires 84,561 synthetic videos for K100-LT ($\fillthresh{=}990$) and 9,195 for UCF-LT ($\fillthresh{=}121$). Generating these required 9.2K and 1.0K GPU hours, respectively, on an H100. For ablations, we set $\fillthresh$=330 for K100-LT, limiting generation to a fixed budget of 2.5K GPU hours.  
We note that this is a one-time offline cost, akin to dataset collection, that incurs zero additional inference latency and will decrease as generative models become faster.

To verify generation quality, we conducted a user study mirroring Kinetics~\cite{kay2017kinetics} data curation process. Annotators were shown five candidate-generated videos along with an action label, then asked to identify which videos depicted the target action (from multiple correct answers). Across 500 trials, users correctly identified the videos 87\% of the time, confirming they are semantically correct and human-recognisable. Full details in Supp~\ref{supp:user-study}. 

\noindent \textbf{Backbone ($f_\theta$)}. We instantiate $f_\theta$ as a VideoMAE (ViT-B) model~\cite{tong2022videomae} pre-trained on Kinetics-400 with the self-supervised objective of video masked auto-encoding. 
This backbone remains the SOTA on action recognition benchmarks when fine-tuned. To reduce the computational burden, we fine-tune only the last encoder layer and the classification head (updating 7.4M$/$86M parameters).

\noindent \textbf{Implementation Details.} 
In Stage~1, we fine-tune for 100 epochs with a base learning rate of $5\cdot10^{-3}$; in Stage~2, for 35 epochs at $5\cdot10^{-4}$. Both stages use AdamW with a weight decay of $0.05$, batch size of 84, cosine learning rate decay with 5-epoch linear warm-up, and an input resolution of $224\times224$, following the standard VideoMAE settings~\cite{tong2022videomae}. With full balancing, training takes 55 hours for K100-LT and 18 hours for UCF-LT on one Nvidia H100 GPU.

\begin{table*}[t]
  \caption{
    \textbf{Long-tail results on K100-LT and UCF-LT.} 
    We compare our method (\textbf{Gen2Balance}) against related long-tail baselines and report class-average accuracy (C/A). \textit{Gen.} marks methods that use generated data; all sharing the same generated data from our pipeline for fair comparison. Full-dataset baseline is shown in gray.
  }
  \vspace*{-8pt}
  \label{tab:main_results}
  \centering
  \scriptsize
  \setlength{\tabcolsep}{4pt}
  
  \begin{tabular}{@{} l c
    >{\columncolor{lt_few}}c >{\columncolor{lt_tail}}c >{\columncolor{lt_head}}c c 
    >{\columncolor{lt_few}}c >{\columncolor{lt_tail}}c >{\columncolor{lt_head}}c c @{}}
    \toprule
    & & \multicolumn{4}{c}{\textbf{K100-LT}} & \multicolumn{4}{c}{\textbf{UCF-LT}} \\
    \cmidrule(lr){3-6} \cmidrule(l){7-10}
    Method & Gen. & Few & Tail & Head & Avg C/A & Few & Tail & Head & Avg C/A \\
    \midrule
    \textcolor{gray}{CE (\textit{Full Dataset})} & \xmark & \textcolor{gray}{74.9} & \textcolor{gray}{81.7} & \textcolor{gray}{88.4} & \textcolor{gray}{80.4} & \textcolor{gray}{92.7} &  \textcolor{gray}{89.4}  & \textcolor{gray}{94.0} & \textcolor{gray}{92.0} \\
    \midrule
    CE & \xmark & 23.4 & 63.3 & \textbf{94.7} & 54.8 & 45.7 & 79.6 & \underline{95.7} & 64.0 \\
    BSCE~\cite{ren2020balanced} & \xmark & 48.7 & 68.5 & 86.7 & 64.5 & 74.3 & 83.0 & 88.5 & 79.3 \\
    cRT~\cite{kangdecoupling} & \xmark & 49.6 & \underline{69.6} & 88.1 & \underline{65.6} & 79.7 & 86.2 & 92.3 & \underline{83.8} \\
    Logit Adj.~\cite{menon2021logit} & \xmark & \underline{55.5}  & 68.4 & 78.5 & \underline{65.6} & \underline{81.2} & 83.4 & 88.6 & 83.1 \\
    LiVT~\cite{xu2023livt} & \xmark & 47.1 & 68.7 & 91.0 & 64.7 & 78.4 & 85.4 & 93.4 & 82.3 \\
    LMR~\cite{perrett2023use} &\xmark & 51.5 & 68.7 & 83.1 & 65.1 & 79.6 & 84.6 & 94.4 & 83.6 \\ 
    EWB-FDR~\cite{hasegawa2024exploring} & \xmark & 47.3 & 69.0 & 89.0 & 64.7 & 78.7 & 86.9 & 92.9 & 83.5 \\
    Sariyildiz~\etal~\cite{sariyildiz2023fake} &\cmark & 26.2 & 58.8 & 93.2 & 52.8 & 73.1 & 86.4 & 94.3 & 80.6 \\
    Li \etal~\cite{li2025role} & \cmark & 23.8 & 62.6 & \underline{93.4} & 54.4 & 75.2 & \underline{88.1} & \textbf{95.9} & 82.5 \\
    \textbf{Gen2Balance} & \cmark & \textbf{62.2} & \textbf{75.0} & 88.4 & \textbf{72.6} & \textbf{86.7} & \textbf{90.3} & 93.4 & \textbf{88.9} \\
    
    \bottomrule
  \end{tabular}
  \vspace{-12pt}
\end{table*}

\subsection{Results}

\noindent \textbf{Long-Tailed Baselines.}
We compare against state-of-the-art (SOTA) and standard long-tail baselines. \textbf{CE} trains with standard cross-entropy and instance-balanced sampling.
\textbf{Balanced Softmax} (\textbf{BSCE})~\cite{ren2020balanced} adjusts logit margins based on class frequency. \textbf{Classifier Retraining}
(\textbf{cRT})~\cite{kangdecoupling} decouples representation from classifier learning: the backbone is trained with CE, then frozen while the classifier head is retrained with class-balanced sampling.
\textbf{Logit Adjustment} (\textbf{Logit Adj.}) applies post-hoc logit correction to CE based on class priors.
\textbf{LiVT}~\cite{xu2023livt} uses bias-corrected cross-entropy with logit adjustment. 
\textbf{LMR}~\cite{perrett2023use} is the SOTA video long-tail baseline that augments in feature space by constructing new samples from class-size-weighted linear combinations of existing features.
\textbf{EWB-FDR}~\cite{hasegawa2024exploring} is the SOTA image long-tail baseline that combines weight balancing with feature diversity regularisation.
Among generative augmentation methods, \textbf{Sariyildiz~\etal}~\cite{sariyildiz2023fake} pretrain on fully balanced synthetic data and fine-tune a new linear head over the frozen backbone, serving as a generative transfer baseline. \textbf{Li~\etal}~\cite{li2025role} follow a sequential strategy: pretrain on synthetic data, then fine-tune on real data with uncertainty-based label smoothing. 
We exclude methods requiring training a class-conditional video generator~\cite{shao2024diffult,hu2025majority} due to prohibitive cost and lack of public implementation. For fair comparison, all methods use the same VideoMAE backbone and fine-tuning protocol, and all generative baselines use the same generated data from our pipeline.

\begin{figure*}[t]
  \centering
 \includegraphics[width=\linewidth]{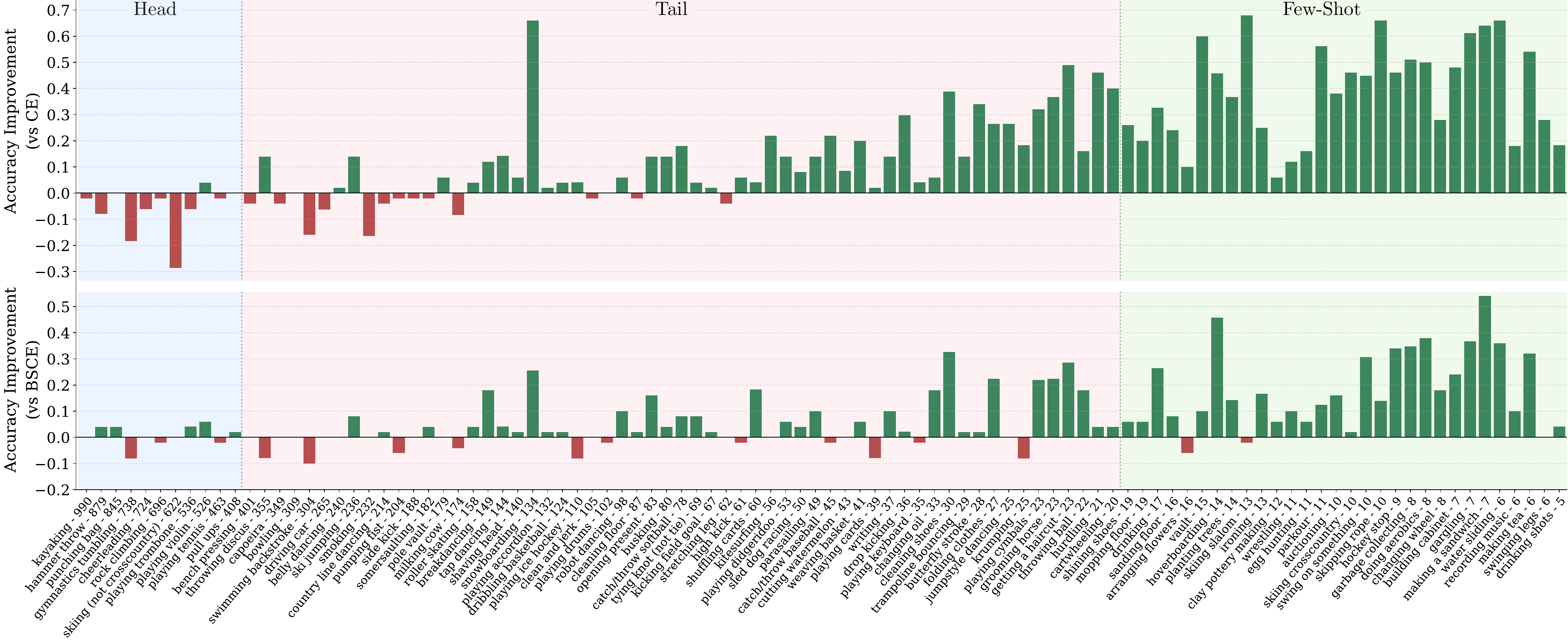}
 \vspace*{-16pt}
  \caption{Per-class accuracy improvement of Gen2Balance over the CE baseline (top) and BSCE~\cite{ren2020balanced} (bottom) on K100-LT. Classes are ordered by size, from largest to smallest (left to right). Background colour indicates head/tail/few-shot splits.
  }
  \label{fig:relative_improvement}
   \vspace{-10pt}
\end{figure*}

\noindent \textbf{Comparative Results.}
We present results in 
Table~\ref{tab:main_results}. 
As expected, the \textbf{CE} baseline is heavily biased towards head classes, achieving the highest head accuracy (94.7\% on K100-LT) but has the lowest overall performance. \textbf{BSCE} and \textbf{cRT} improve few-shot class performance, validating algorithmic interventions for class imbalance. 
\textbf{Logit Adj.} achieves the strongest few-shot performance among non-generative methods (55.5\% on K100-LT), but degrades in head accuracy (78.5\%), reflecting the inherent head–tail trade-off in post-hoc correction.
Its overall performance is comparable to 
\textbf{LMR},
\textbf{LiVT} and \textbf{EWB-FDR}, with LiVT retaining stronger head performance while EWB-FDR and LMR achieving a more balanced distribution of gains. 
Both generative baselines underperform long-tail baselines, despite access to the same synthetic data as Gen2Balance.
\textbf{Gen2Balance} achieves the highest overall accuracy on both benchmarks (72.6\% on K100-LT, 88.9\% on UCF-LT), surpassing the strongest baselines by +7.0\% and +5.1\%, respectively. Gains are concentrated in tail and few-shot classes, while head accuracy remains competitive with the CE baseline.
Fig.~\ref{fig:relative_improvement} shows per-class improvements against CE and BSCE.

\begin{wraptable}{r}{0.5\textwidth}
  \vspace{-28pt}
  \caption{\textbf{Compositionally Rare Actions (RareAct~\cite{miech2020rareact}).} 22 rare actions appended to K100-LT as few-shot classes. \colorbox{lt_rareact}{RareAct}: Avg C/A on these classes; others refer to the original K100-LT split.}
  \label{tab:rareact}
  \centering
  \scriptsize
  \begin{tabular}{@{} l
    >{\columncolor{lt_rareact}}c 
    >{\columncolor{lt_few}}c 
    >{\columncolor{lt_tail}}c 
    >{\columncolor{lt_head}}c 
    c @{}}
    \toprule
    Method & RareAct & Few & Tail & Head & Avg C/A \\
    \midrule
    CE & 11.3 & 22.9 & 62.6 & \textbf{94.7} & 46.7 \\
    BSCE~\cite{ren2020balanced} & 23.2 & 47.3 & 67.6 & 82.4 & 56.0 \\
    Logit Adj.~\cite{menon2021logit}& 27.8 & 53.1 & 66.3 & 74.1 & 56.8 \\
    \textbf{Gen2Balance} & \textbf{59.7} & \textbf{59.3} & \textbf{72.4} & 86.4 & \textbf{68.2} \\
    \bottomrule
  \end{tabular}
  \vspace{-10pt}
\end{wraptable}

\sethlcolor{lt_rareact}

\noindent \textbf{Rare classes.} To test whether Gen2Balance extends beyond curated long-tail splits, we append 
\hl{22 rare action classes from RareAct}~\cite{miech2020rareact} to K100-LT as few-shot classes, resulting in a 122-class dataset (see Sec.~\ref{sec:experiments} for details). We generate synthetic data using our pipeline and train on the combined dataset (see Table~\ref{tab:rareact}).
We observe that standard baselines struggle: CE achieves only 11.3\% and Logit Adj. 27.8\% on few-shot classes. Gen2Balance reaches 59.7\%, matching the performance of the original K100-LT few-shot classes, confirming that generative filling is effective even for actions that are genuinely rare in both our data and the real world.

\begin{table}[t]
  \caption{
    \textbf{Analysis of Augmented Data Sources.} 
    We compare our generative approach against other data sources on K100-LT $(\fillthresh{=}330)$.
  }
  \label{tab:data_sources}
  \centering
  \scriptsize
  \setlength{\tabcolsep}{6pt}
  \begin{tabular}{@{} l l
    >{\columncolor{lt_few}}c 
    >{\columncolor{lt_tail}}c 
    >{\columncolor{lt_head}}c 
    c @{}}
    \toprule
    Augmented Data Source ($\mathcal{D}_{gen}$)& Data Type & Few & Tail & Head & Avg C/A \\
    \midrule
    None & -  & 48.7 & 68.5 & 86.7 & 64.5 \\
    Web-Retrieval (String Match) & Real Videos  & 50.1 & 70.8 & \textbf{90.4} & 66.7\\
    Web-Retrieval (Semantic Match) & Real Videos & 53.1 & 69.8 & 88.8 & 66.9 \\
    Qwen-Image~\cite{wu2025qwen} (Our Prompting) & Gen. Images & 53.2 & 71.4 & 90.3 & 68.0\\
    WAN~\cite{wan2025wan} (Naive prompting) & Gen. Videos & 49.3 & 68.5 & 89.5 & 65.1 \\
    \textbf{Gen2Balance (Ours)} & Gen. Videos & \textbf{62.1} & \textbf{72.3} & 87.9 & \textbf{70.9} \\
    \bottomrule
  \end{tabular}
 \vspace{-12pt}
\end{table}

\noindent \textbf{Alternative Data Sources.}
We additionally compare against alternative data sources for augmentation on K100-LT (with $\fillthresh$=330), keeping the Gen2Balance training strategy constant (Table~\ref{tab:data_sources}).
\textbf{(i) Web-}\textbf{retrieval.} We test whether retrieving real videos from the internet to augment training outperforms generation, by querying WebVid10M~\cite{bain2021frozen}, a 10M-pair web video-caption dataset commonly used to train video generators. We retrieve clips using two matching strategies: exact string matching and semantic embedding similarity via Qwen-Embedding~\cite{zhang2025qwen3}.
\textbf{(ii) Qwen-Image.}~\cite{wu2025qwen} To isolate whether motion cues are necessary, we generate static images using Qwen-Image conditioned on the same text prompts produced by our pipeline.
\textbf{(iii) WAN (Naive Prompting).} Holding the video generator fixed, we replace our pipeline with templated prompting~\cite{sariyildiz2023fake,he2023is} (\textit{\ie, ``A video depicting the action of [class name]''}). Importantly, to ensure a fair comparison, we maintain the same augmented data volume ($N'_c$) across all data sources.

As shown in Table~\ref{tab:data_sources}, web-retrieval underperforms our generative approach, due to inherent noise in web videos and the true long-tail problem---a tail class may also be scarce or mislabelled on the web (\eg, \textit{Krumping}).
Interestingly, web retrieval is outperformed by using a generative image backbone for tail, few-shot, and overall class performance. However, the generative image backbone still lags behind our generative video approach, confirming that images alone are insufficient to differentiate dynamic actions, which require motion cues (\eg, differentiating \textit{Breakdancing} from \textit{Krumping}). Naive prompting underperforms all these baselines, producing repetitive samples that lack diversity and struggle with ambiguity. Gen2Balance achieves the best overall performance.

\begin{figure*}[t]
  \centering
 \includegraphics[width=\linewidth]{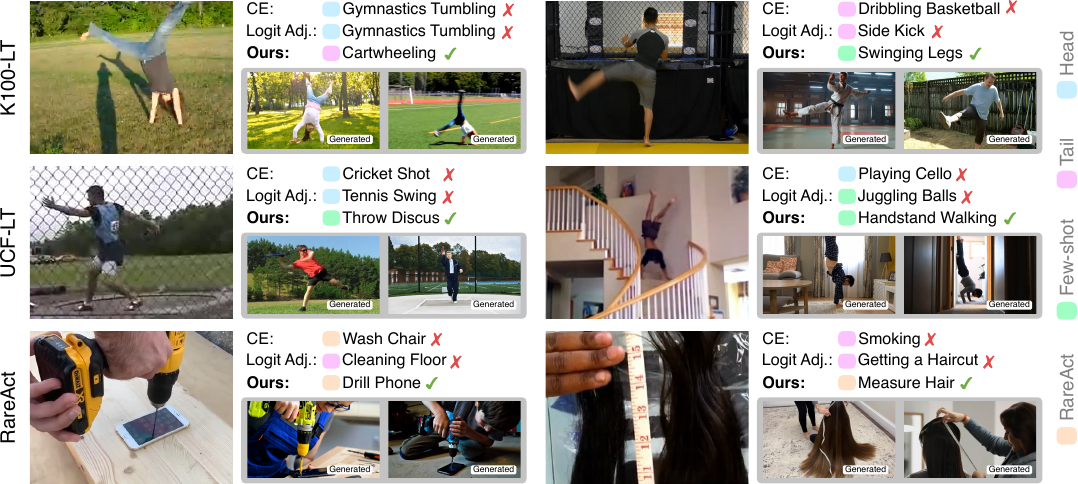}
  \caption{\textbf{Qualitative Results} from three datasets comparing CE, Logit Adj., and Gen2Balance (Ours). Colours indicate the category of the predicted class. For each test video, we also show the two nearest generated samples from $\mathcal{D}_{gen}$ in feature space.}
  \label{fig:qualitative}
  \vspace{-3pt}
\end{figure*}

\paragraphcustom{Qualitative Results}. \cref{fig:qualitative} shows test videos across all benchmarks where CE and Logit Adj.\ fail but Gen2Balance succeeds. For each sample, we show the two nearest neighbours from $\mathcal{D}_{gen}$ in the feature space of Gen2Balance. The generated videos match the test sample in environment, viewpoint, and actor appearance.
For example, the top example shows cartwheeling in an outdoor scene, and its nearest generated videos depict the same action in a similar setting and camera angle.
These results are consistent across actions and datasets.

\subsection{Ablations}

We ablate key components of Gen2Balance: the training strategy, the generation pipeline, and the amount of synthetic data used.

\begin{table}[tb]
  \caption{
    \textbf{Ablation of Gen2Balance Training Strategy.} 
    We analyse the impact of generative data ($\mathcal{D}_{gen}$), loss functions, real ($N$) vs aug. ($N+N'$) frequency sources used in the BSCE loss, and the two-stage training on K100-LT ($\fillthresh=330$). 
    \textit{Epochs} denotes training duration (Stage 1, Stage 2).
    Row (g) (Ours) achieves the best trade-off.
  }
  \label{tab:ablation_method}
  \centering
  \scriptsize 
  \setlength{\tabcolsep}{3pt}
  
  \begin{tabular}{r l l l l c >{\columncolor{lt_few}}c >{\columncolor{lt_tail}}c >{\columncolor{lt_head}}c c}
    \toprule
    & Training Data & Loss & Freq. Src. & Epochs & Stage 2 & Few & Tail & Head & Avg C/A\\
    \midrule
    (a) & $\mathcal{D}_{train}$ & CE & - & 100 & \ding{55} & 23.4 & 63.3 & \textbf{94.7} & 54.8 \\
    (b) & $\mathcal{D}_{train} \cup \mathcal{D}_{gen}$ & CE & - & 100 & \ding{55} & 27.3 & 55.2 & 91.2 & 50.8 \\
    \midrule
    (c) & $\mathcal{D}_{train} \cup \mathcal{D}_{gen}$ & BSCE & $N+N'$ & 100 & \ding{55} & 41.8 & 65.7 & 90.5 & 61.3 \\
    (d) & $\mathcal{D}_{train} \cup \mathcal{D}_{gen}$ & BSCE & $N+N'$ & 100, 35 & \ding{51} & 61.0 & \textbf{72.4} & 87.7 & 70.6 \\
    \midrule
    (e) & $\mathcal{D}_{train} \cup \mathcal{D}_{gen}$ & BSCE & $N$ & 100 & \ding{55} & 61.9 & 68.3 & 82.7 & 67.8 \\
    (f) & $\mathcal{D}_{train} \cup \mathcal{D}_{gen}$ & BSCE & $N$ & 135 & \ding{55} & 54.6 & 70.7 & 89.7 & 67.9 \\
    \textbf{(g)} & \textbf{$\mathcal{D}_{train} \cup \mathcal{D}_{gen}$} & \textbf{BSCE} & \textbf{$N$} & \textbf{100, 35} & \textbf{\ding{51}} & \textbf{62.1} & 72.3 & 87.9 & \textbf{70.9} \\
  \bottomrule
  \end{tabular}
   \vspace{-12pt}
\end{table}

\paragraphcustom{Training Strategy}. Table~\ref{tab:ablation_method} dissects our training strategy on K100-LT. Comparing \textbf{(a)} vs. \textbf{(b)} reveals that simply adding generated data with standard CE loss degrades the accuracy ($-4$\%). 
Using a class-balanced loss (BSCE) in \textbf{(c)} stabilises model training. Comparing \textbf{(c)} vs. \textbf{(d)} shows inclusion of Stage 2 (fine-tuning on real data) boosts performance on all metrics, possibly rectifying the domain gap introduced by the generated data in Stage 1 \textbf{(c)}. Comparing \textbf{(d)} vs. \textbf{(g)} shows that calculating BSCE loss margins solely based on frequencies of real video clips rather than total augmented frequency creates a \textit{super-margin} effect for the minority classes, yielding better few-shot accuracy (62.1\% vs 61.0\%) with retention of the head performance. Finally, row \textbf{(f)} trains Stage 1 for 135 epochs (matching the iterations in \textbf{(g)}). It fails to match the performance of \textbf{(g)}, confirming our gains do not come from extended training.

\begin{table}[tb]
  \caption{
    \textbf{Ablation of the Generative Pipeline.}
    We ablate pipeline components on 10 randomly selected tail and few-shot classes from K100-LT, modifying only their synthetic data while fixing all other classes at the full pipeline~(e).
    \textit{\# Cls Modified}: number of modified classes.
    \textit{FVD}: per-class Fr\'{e}chet Video Distance~\cite{unterthiner2019fvd} against real Kinetics-100 videos ($\downarrow$\,=\,lower is better).
    \textit{ViCLIP}: per-class ViCLIP semantic similarity~\cite{wang2024internvid} between generated videos and the prompt ``\textit{a video depicting the action of [class name]}''.
    \textit{Avg}\,/\,\textit{Overall Avg}: accuracy on the 10 ablated classes\,/\,all 100 classes.
  }
  \label{tab:pipeline_ablation}
  \centering
  \scriptsize
  \setlength{\tabcolsep}{3pt}
\begin{tabular}{@{} c l c c c c c  @{}}
    \toprule
    & Pipeline Stage &\# Cls Modified & FVD$\downarrow$ & ViCLIP$\uparrow$ & Avg$\uparrow$ & \textcolor{gray}{Overall Avg$\uparrow$} \\
    \midrule
    (a) & Naive Prompting (all classes) &100 & 1277.1 & 0.1635 &37.3 &\textcolor{gray}{65.1}\\
    (b) & Naive Prompting &10 & 1277.1 & 0.1635 & 37.3 & \textcolor{gray}{68.9} \\
    (c) & + Diversified Prompts & 10 & 886.8 & 0.1585 & 44.9 & \textcolor{gray}{69.5} \\
    (d) & + Action Profile &10 & 842.8 & 0.1649 & 47.2 & \textcolor{gray}{70.0} \\
    (e) & + In-Context Exemplars (Ours) &0 & \textbf{836.1} & \textbf{0.1774} & \textbf{49.8} & \textcolor{gray}{\textbf{70.9}} \\
    \bottomrule
  \end{tabular}
\end{table}

\paragraphcustom{Generation Pipeline.}
Table~\ref{tab:pipeline_ablation} ablates our generation pipeline. As generating all 100 classes is expensive, we randomly select 10 tail and few-shot classes and ablate only their synthetic data, fixing all others (reducing cost from 2.5K to 330 GPU hours at $\fillthresh{=}330$). Alongside accuracy on these 10 classes and overall, we report two generation-quality metrics: FVD~\cite{unterthiner2019fvd}, measuring distributional alignment with real Kinetics-100 videos, and ViCLIP~\cite{wang2024internvid}, measuring semantic similarity to the class label. Rows \textbf{(a)}--\textbf{(b)} isolate the impact of the 10 selected classes by applying naive prompting to all 100 vs.\ only these 10. Each subsequent stage \textbf{(b)}--\textbf{(e)} consistently improves accuracy and FVD.

\begin{figure*}[t]
  \centering
  \includegraphics[width=\linewidth]{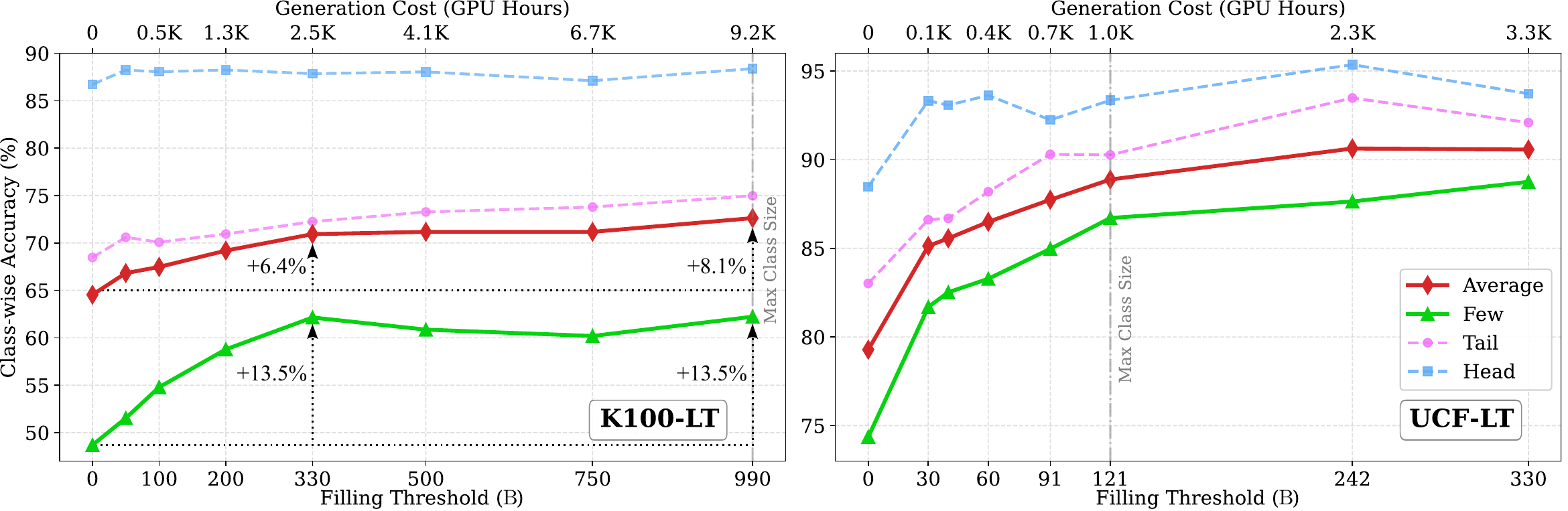}
  \caption{\textbf{Effect of Scaling Generated Data}. We analyse the impact of Filling Threshold $\fillthresh$ (bottom axis) and corresponding generation cost in GPU hours (top axis) on model performance. \textbf{(Left)} On K100-LT, while full balancing ($\fillthresh{=}990$) yields the highest accuracy, partial balancing ($\fillthresh{=}330$) achieves competitive results at just $27\%$ of the compute cost. \textbf{(Right)} On UCF-LT, performance continues to improve even when oversampling beyond the maximum head class size ($\fillthresh{>}121$), showing that our data augmentation also addresses general data scarcity in smaller datasets.}
  \vspace*{-12pt}
  \label{fig:scaling_generated_data}
\end{figure*}

\paragraphcustom{Scaling Generated Data.} Unlike real data collection, generative augmentation is unbounded by size. Hence, in Fig.~\ref{fig:scaling_generated_data}, we investigate the trade-off between performance and computational cost (measured in GPU hours) as we vary the filling threshold $\fillthresh$. Note that $\fillthresh{=}0$ represents the BSCE baseline, while all runs with $\fillthresh>0$ follow the same Gen2Balance training strategy. (i) In K100-LT, which is the larger-scale dataset, we observe diminishing returns. The best performance is achieved with full balancing ($\fillthresh{=}990$), boosting Avg C/A to \textbf{72.6\%} and few-shot to \textbf{62.2\%}. However, this requires relatively large compute (9.2K GPU hours). Notably, a partial balancing strategy ($\fillthresh{=}330$) already achieves clear gains (Avg 70.9\%) using only 2.5K GPU hours (\ie, $79\%$ of the gains over BSCE at only $27\%$ of the compute cost).
(ii) In UCF-LT, which is the smaller-scale dataset, the plot also shows a sharp increase with $\fillthresh$. Unlike K100-LT, performance continues to improve even when we oversample beyond the maximum class size ($\fillthresh{>}121$). 
Performance on few-shot classes massively increases from  74.4\% to 88.8\% (+14.4\%).
This suggests that, for smaller-scale benchmarks, generative filling not only corrects the long-tail imbalance but also serves as a form of data augmentation that may benefit the entire distribution.

\begin{wraptable}{r}{0.5\textwidth}
  \vspace{-32pt}
  \caption{
    \textbf{Robustness to Web-Popularity Re-indexing.}
    Performance on the original K100-LT split vs.\ the
    Web-Popularity re-indexed split, where few-shot classes
    are statistically rare on the web.
  }
  \label{tab:webvid_split}
  \centering
  \scriptsize
  \setlength{\tabcolsep}{3pt}
  \begin{tabular}{@{} l
    >{\columncolor{lt_few}}c 
    >{\columncolor{lt_tail}}c 
    >{\columncolor{lt_head}}c 
    c @{}}
    \toprule
    Method & Few & Tail & Head & Avg C/A \\
    \midrule
    \multicolumn{5}{l}{\textit{Original K100-LT Split}} \\
    CE & 23.4 & 63.3 & \textbf{94.7} & 54.8 \\
    BSCE~\cite{ren2020balanced} & 48.7 & 68.5 & 86.7 & 64.5 \\
    Logit Adj.~\cite{menon2021logit} & 55.5 & 68.4 & 78.5 & 65.6 \\
    \textbf{Gen2Balance} & \textbf{62.1} & \textbf{72.3} & 87.9 & \textbf{70.9} \\
    \midrule
    \multicolumn{5}{l}{\textit{Web-Popularity K100-LT Split}} \\
    CE & 21.7 & 64.1 & \textbf{87.8} & 54.0 \\
    BSCE~\cite{ren2020balanced} & 49.4 & 66.8 & 79.2 & 63.0 \\
    Logit Adj.~\cite{menon2021logit} & 52.9 & 68.6 & 69.9 & 64.0 \\
    \textbf{Gen2Balance} & \textbf{71.9} & \textbf{72.6} & 80.5 & \textbf{73.2} \\
    \bottomrule
  \end{tabular}
  \vspace{-12pt}
\end{wraptable}

\paragraphcustom{Robustness to Generative Priors} (\textbf{Web-Reindexing}). 
To ensure a rigorous evaluation, we account for the possibility that certain few-shot classes are already well-represented in the training data of
$\mathcal{G}$, making them easy to generate. 
We re-index K100-LT classes by their frequency in WebVid10M~\cite{bain2021frozen}, using web popularity as a proxy for $\mathcal{G}$'s training distribution. This places genuinely scarce actions as few-shot (\eg, \textit{Playing Didgeridoo}, \textit{Krumping}, and \textit{Jumpstyle Dancing}).

As shown in Table~\ref{tab:webvid_split}, Gen2Balance still outperforms Logit Adj. by +19\% on few-shot classes, with overall accuracy comparable to standard K100-LT ordering. This confirms our approach is robust when the generator has potentially limited exposure to the target actions.

\vspace{-6pt}
\section{Conclusion}

We address long-tailed video action recognition through generative balancing. Our approach combines an LLM-driven prompt pipeline leveraging diversity criteria, action profiles, and in-context exemplars with a two-stage training strategy over real and synthetic data. On long-tailed versions of UCF-101 and Kinetics, Gen2Balance surpasses state-of-the-art long-tail baselines, and our generated data outperforms alternative data sources. We further find that: (1) Gen2Balance is robust to scarce actions with rare verb-noun compositions, and (2) a partial balancing strategy captures the majority of performance gains at a fraction of the generation cost. We publicly release our 140K generated videos to support future research on generative balancing for long-tailed video understanding.

\section*{Acknowledgements}
{\small
This work was supported by EPSRC Fellowship UMPIRE (EP/T004991/1) and a charitable donation from Adobe to the University of Bristol. We acknowledge the usage of GPU Node hours granted as part of the AIRR Gateway project ``HOI Foundational Model from Egocentric Data'' (Dec 2025--Mar 2026) and the Sovereign AI Unit call project ``Gen Model in Ego-sensed World'' (Aug 2025--Nov 2025).
}

\bibliographystyle{splncs04}
\bibliography{main}

\appendix

\section*{Supplementary Material for Gen2Balance: Generative Balancing for Long-Tailed Video Action Recognition}
\label{sec:appendix}

\makeatletter
{%
\newcommand{\tocentry}[3]{%
  \par\noindent
  \setlength{\rightskip}{1.5em}%
  \setlength{\parfillskip}{-\rightskip}%
  \textbf{#1}\hspace{6pt}#3%
  \nobreak
  \leaders\hbox{$\m@th\mkern 4.5mu\hbox{.}\mkern 4.5mu$}\hfill
  \nobreak
  \hb@xt@1.5em{\hss\pageref{#2}}\par
  \vspace{2pt}%
}%
\newcommand{\tocsubentry}[3]{%
  \par\noindent
  \setlength{\rightskip}{1.5em}%
  \setlength{\parfillskip}{-\rightskip}%
  \hspace{1.8em}\textbf{#1}\hspace{6pt}#3%
  \nobreak
  \leaders\hbox{$\m@th\mkern 4.5mu\hbox{.}\mkern 4.5mu$}\hfill
  \nobreak
  \hb@xt@1.5em{\hss\pageref{#2}}\par
  \vspace{2pt}%
}%
\vspace{4pt}%
\noindent\textbf{Table of Contents}\\
\vspace{-4pt}
\noindent\rule{\linewidth}{0.5pt}%
\vspace{6pt}%
\tocentry{A}{supp:showcase-video}{Showcase Video}
\tocentry{B}{supp:additional-expts}{Additional Experiments}
\tocsubentry{B.1}{subsec:frozen_vs_full_finetune}{Frozen Layers vs Full Fine-Tuning of the Backbone}
\tocsubentry{B.2}{supp:backbone_generalisation}{Backbone Generalisability}
\tocsubentry{B.3}{supp:vlm_generalisability}{VLM Generalisability}
\tocsubentry{B.4}{supp:filling_to_orig}{Filling Gen2Balance to Original Data Size}
\tocsubentry{B.5}{supp:long_tail_baselines_with_our_gen}{Training Long-Tailed Baselines with our $\mathcal{D}_{gen}$ Data}
\tocsubentry{B.6}{supp:wan_test_leakage}{Evaluating Test-Set Memorisation in WAN}
\tocentry{C}{sec:dataset_details}{Dataset Details}
\tocsubentry{C.1}{supp:long-tail-creation}{Creation of Long-Tail Action Recognition Datasets}
\tocsubentry{C.2}{supp:rareact}{RareAct Classes Selection}
\tocentry{D}{supp:user-study}{User Study of Generated Videos}
\tocentry{E}{sec:supp_prompts}{Gen2Balance LLM Prompts}
\tocsubentry{E.1}{sec:supp_prompt_profile}{Action Profile Generation Prompt}
\tocsubentry{E.2}{sec:supp_prompt_diversity}{Diverse Text-to-Video Prompt Generation}
\tocentry{F}{supp:action-profiles}{Sample Action Profiles}
\tocentry{G}{supp:text-prompts}{Sample Text Prompts}
\noindent\rule{\linewidth}{0.5pt}%
\vspace{8pt}%
}
\makeatother

\section{Showcase Video}
\label{supp:showcase-video}

To demonstrate the quality of our generated videos, we showcase randomly sampled generated videos from 10 classes across K100, UCF, and RareAct, which can be viewed in this link: \url{https://prajwalgatti.github.io/gen2balance/showcase.mp4}.

\begin{table*}[h]
    \caption{\textbf{Frozen layers vs.\ full fine-tuning of VideoMAE.} Full fine-tuning updates all 86M parameters of VideoMAE, and frozen layers update only the last encoder layer and classification head (7.4M). Gen2Balance outperforms all baselines in both regimes. \textcolor{gray}{Grey} denotes the upper bound trained on the full dataset.}
  \label{tab:full_vs_partial_training}
  \centering
  \scriptsize
  \setlength{\tabcolsep}{4pt}
  
  \begin{tabular}{@{} l c 
    >{\columncolor{lt_few}}c >{\columncolor{lt_tail}}c >{\columncolor{lt_head}}c c 
    >{\columncolor{lt_few}}c >{\columncolor{lt_tail}}c >{\columncolor{lt_head}}c c @{}}
    \toprule
    && \multicolumn{4}{c}{\textbf{K100-LT}} & \multicolumn{4}{c}{\textbf{UCF-LT}} \\
    \cmidrule(lr){3-6} \cmidrule(l){7-10}
    Method & Gen. & Few & Tail & Head & Avg C/A & Few & Tail & Head & Avg C/A \\
    \midrule
    \multicolumn{9}{@{}l}{\textit{Full fine-tuning}} \\
    \addlinespace[2pt]
    \textcolor{gray}{CE (\textit{Full Dataset})} &\textcolor{gray}{\xmark} & \textcolor{gray}{81.8} & \textcolor{gray}{87.4} & \textcolor{gray}{93.2} & \textcolor{gray}{83.4} & \textcolor{gray}{96.5} & \textcolor{gray}{92.3} & \textcolor{gray}{96.7} & \textcolor{gray}{95.4} \\
    CE & \xmark & 30.8 & 68.8 & \underline{95.4} & 60.3 & 79.2 & 90.8 & \underline{96.6} & 85.5 \\
    cRT~\cite{kangdecoupling} & \xmark & 49.1 & \underline{73.6} & 93.2 & 68.4 & 82.6 & 90.1 & 95.9 & 87.1 \\
    Logit Adj.~\cite{menon2021logit} & \xmark & \underline{57.9} & 73.2 & 88.6 & \underline{70.3} & \underline{85.8} & 89.0 & 95.9 & 88.5 \\
    Li~\etal~\cite{li2025role} & \cmark & 27.0 & 66.2 & \textbf{96.3} & 57.7 & 85.5 & \textbf{93.6} & \textbf{98.5} & \underline{90.1} \\
    \textbf{Gen2Balance} & \cmark & \textbf{65.5} & \textbf{79.3} & 94.4 & \textbf{78.7} & \textbf{90.9} & \underline{93.5} & 95.3 & \textbf{92.4} \\
    \midrule
    \multicolumn{9}{@{}l}{\textit{Frozen layers}} \\
    \addlinespace[2pt]
    \textcolor{gray}{CE (\textit{Full Dataset})} & \textcolor{gray}{\xmark} & \textcolor{gray}{74.9} & \textcolor{gray}{81.7} & \textcolor{gray}{88.4} & \textcolor{gray}{80.4} & \textcolor{gray}{92.7} & \textcolor{gray}{89.4} & \textcolor{gray}{94.0} & \textcolor{gray}{92.0} \\
    CE & \xmark & 23.4 & 63.3 & \textbf{94.7} & 54.8 & 45.7 & 79.6 & \underline{95.7} & 64.0 \\
    cRT~\cite{kangdecoupling} & \xmark & 49.6 & \underline{69.6} & 88.1 & 65.6 & 79.7 & 86.2 & 92.3 & \underline{83.8} \\
    Logit Adj.~\cite{menon2021logit} & \xmark & \underline{55.5}  & 68.4 & 78.5 & \underline{65.6} & \underline{81.2} & 83.4 & 88.6 & 83.1 \\
    Li~\etal~\cite{li2025role} & \cmark & 23.8 & 62.6 & \underline{93.4} & 54.4 & 75.2 & \underline{88.1} & \textbf{95.9} & 82.5 \\
    \textbf{Gen2Balance} & \cmark & \textbf{62.2} & \textbf{75.0} & 88.4 & \textbf{72.6} & \textbf{86.7} & \textbf{90.3} & 93.4 & \textbf{88.9} \\
    \bottomrule
  \end{tabular}
\end{table*}

\section{Additional Experiments}
\label{supp:additional-expts}
\subsection{Frozen Layers vs Full Fine-Tuning of the Backbone}
\label{subsec:frozen_vs_full_finetune}
Table~\ref{tab:full_vs_partial_training} compares two fine-tuning settings for VideoMAE: full fine-tuning (all 86M parameters) against our default frozen-layer setting (\ie, updating only the last encoder layer and classification head, 7.4M parameters). 
Both Gen2Balance settings use fully balanced generation, \ie, $B=990$ for K100-LT and $B=121$ for UCF-LT.

Fine-tuning consistently improves accuracy by a small (fairly fixed) margin of typically 3–6\%, with CE on the small UCF-LT being a notable exception.
Gen2Balance achieves accuracy of 78.7\% on K100-LT and 92.4\% on UCF-LT. Importantly, the relative ranking of methods remains consistent across both settings where Gen2Balance outperforms all baselines in both regimes, and the gains over the strongest non-generative baseline (Logit Adj.) remain substantial (+8.4\% with full fine-tuning vs.\ +7.0\% with frozen layers on K100-LT). 

We adopt the frozen-layer setting throughout the main paper as it is more computationally efficient.

\begin{table*}[t]
  \caption{\textbf{Backbone Generalisability of Gen2Balance}. We evaluate the Gen2Balance strategy on K100-LT with two distinct video backbones: VideoMAE and V-JEPA 2, using a filling threshold of $B=330$. Gen2Balance consistently improves over the baselines across both backbone architectures and pretraining paradigms. Pre-train hrs.\ and Gen.\ hrs.\ report approximate compute (H100 GPU-hours) for backbone pre-training and synthetic-data generation.
  }
  \label{tab:backbone_generalizability}
  \centering
  \scriptsize
  \setlength{\tabcolsep}{4pt}
  
  \begin{tabular}{@{} l c c c
    >{\columncolor{lt_few}}c >{\columncolor{lt_tail}}c >{\columncolor{lt_head}}c c @{}}
    \toprule
    Method & Gen. & Pre-train hrs. & Gen. hrs. & Few & Tail & Head & Avg C/A \\
    \midrule
    \multicolumn{7}{@{}l}{\textit{VideoMAE backbone}~\cite{tong2022videomae}} \\
    \addlinespace[2pt]
    CE & \xmark & $\sim$0.8K & 0 & 23.4 & 63.3 & \textbf{94.7} & 54.8 \\
    BSCE~\cite{ren2020balanced} & \xmark & $\sim$0.8K & 0 & 48.7 & \underline{68.5} & 86.7 & 64.5 \\
    Logit Adj.~\cite{menon2021logit} & \xmark & $\sim$0.8K & 0 & \underline{55.5}  & 68.4 & 78.5 & \underline{65.6} \\
    \textbf{Gen2Balance} & \cmark & $\sim$0.8K & 2.5K & \textbf{62.1} & \textbf{72.3} & \underline{87.9} & \textbf{70.9} \\
    \midrule
    \multicolumn{7}{@{}l}{\textit{V-JEPA 2 backbone}~\cite{assran2025vjepa2}} \\
    \addlinespace[2pt]
    CE & \xmark & $\sim$7.5K & 0 & 46.6 & 71.1 & \textbf{92.8} & 66.1 \\
    BSCE~\cite{ren2020balanced} & \xmark & $\sim$7.5K & 0 & 59.3 & \underline{74.2} & 88.2 & \underline{71.3} \\
    Logit Adj.~\cite{menon2021logit} & \xmark & $\sim$7.5K & 0 & \underline{61.3} & 71.7 & 86.4 & 70.2 \\
    \textbf{Gen2Balance} & \cmark & $\sim$7.5K & 2.5K & \textbf{68.9} & \textbf{77.3} & \underline{90.8} & \textbf{76.3} \\
    \bottomrule
  \end{tabular}
\end{table*}

\subsection{Backbone Generalisability}
\label{supp:backbone_generalisation}
To verify whether the Gen2Balance strategy generalises to an alternative video backbone ($f_\theta$), we also evaluate using the V-JEPA 2~\cite{assran2025vjepa2} pre-trained model. Compared to the 86M-parameter VideoMAE (base-variant) model used in the main text, this larger 375M-parameter (large-variant) backbone employs a distinct joint-embedding predictive architecture. For a fair comparison, we use the same fine-tuning procedure, updating only the last layer, the pooling layer, and the classification head of V-JEPA 2. We keep the filling threshold as $B=330$, aligned with our other ablations. 

As shown in Table~\ref{tab:backbone_generalizability}, the higher-capacity V-JEPA 2 naturally improves all baselines and results. The CE baseline accuracy increases to 66.1\%. However, Gen2Balance remains the strongest method, achieving 76.3\% accuracy, with a +10.2\% improvement over CE and a +5.0\% improvement over the strongest long-tailed baseline (BSCE). Notably, few-shot accuracy improves the most (from 46.6\% to 68.9\%), which can be attributed to increased capacity and a larger pre-training dataset. These results confirm that Gen2Balance contributes significantly to different backbones and scales effectively alongside larger, more powerful backbones.

Since V-JEPA 2, a larger backbone, improves all methods, a natural question is whether the generation budget is better spent on pre-training a larger backbone. Table~\ref{tab:backbone_generalizability} assesses this by comparing performance to approximate H100 GPU-hour budgets\footnote{The V-JEPA 2 pre-training cost is estimated from~\cite{assran2025vjepa2}, which does not directly report GPU-hours in H100 units.}. With the \emph{smaller} ViT-B VideoMAE, adding generated data (2.5K gen-hours at $B{=}330$) reaches 70.9\% Avg C/A, surpassing the larger ViT-L V-JEPA 2 fine-tuned on real data alone (66.1\%); despite VideoMAE being much cheaper to pre-train ($\sim$0.8K vs.\ $\sim$7.5K hours). Generation is thus a more effective use of compute than scaling the backbone, and the two are also complementary: V-JEPA 2 with Gen2Balance achieves the best accuracy (76.3\%) albeit at a higher total cost ($\sim$10K hours).

\subsection{VLM Generalisability}
\label{supp:vlm_generalisability}
To test whether the Gen2Balance strategy generalises to an alternative VLM ($\mathcal{M}$), we replace Gemini 2.5 Pro with the open-source Qwen3-VL-32B~\cite{bai2025qwen3} and re-run the full pipeline on the same 10 tail and few-shot classes from K100-LT as Table~\ref{tab:pipeline_ablation} ($B=330$). We concatenate the in-context exemplars into a single long clip before conditioning Qwen3-VL to generate action profiles. Qwen3-VL prompts yield 49.2\% average accuracy on these 10 classes, comparable to the 49.8\% with Gemini, confirming that Gen2Balance generalises across~$\mathcal{M}$.

Both models also produce comparably specific, diverse, and class-faithful prompts. For \textit{robot dancing}, Qwen3-VL generates \textit{``A solo dancer in a black and white outfit is robot dancing, moving with sharp, angular motions and freezing mid-step, in an abandoned warehouse with flickering fluorescent lights''}, while Gemini generates \textit{``A street performer is robot dancing on a busy sidewalk, executing sharp, staccato arm movements and isolating his chest to the beat from a nearby boombox''}, both correctly capturing the human-imitating-a-robot interpretation. Quantitatively, embedding each prompt~\cite{zhang2025qwen3}, the cross-$\mathcal{M}$ within-class similarity is $0.69\pm0.05$, on par with within-$\mathcal{M}$ similarity (Qwen $0.71\pm0.06$, Gemini $0.70\pm0.06$) and well above the $0.43\pm0.05$ across-class control.

\begin{table*}[h]
  \caption{\textbf{Balancing Gen2Balance to the original Kinetics-100 class sizes}. Instead of a fixed uniform filling threshold $B$, we generate synthetic videos for each class to match the original full Kinetics-100 training set size.
  }
  \label{tab:filling_to_orig_size}
  \centering
  \scriptsize
  \setlength{\tabcolsep}{4pt}
  \begin{tabular}{@{} l 
    >{\columncolor{lt_few}}c >{\columncolor{lt_tail}}c >{\columncolor{lt_head}}c c @{}}
    \toprule
    Method & Few & Tail & Head & Avg C/A \\
    \midrule
    CE (Full Dataset) & 74.9 & 81.7 & 88.4 & 80.4 \\
    \textbf{Gen2Balance} (per-class fill) & 59.2 & 72.7 & 89.3 & 70.5 \\
    \textbf{Gen2Balance} (balanced fill) &62.2 &75.0 &88.4 &72.6\\
    \bottomrule
  \end{tabular}
\end{table*}

\subsection{Filling Gen2Balance to Original Data Size}
\label{supp:filling_to_orig}
The original Kinetics-100 distribution is not uniformly class-balanced (its largest class size is 990, and its smallest is 252). For direct comparison, we also add a version of Gen2Balance trained by filling synthetic videos only up to the per-class counts in the dataset.
As shown in Table~\ref{tab:filling_to_orig_size}, the performance remains comparable with a clear advantage to the few-shot classes when additional data is ingested.

\subsection{Training Long-Tailed Baselines with our $\mathcal{D}_{gen}$ Data}
\label{supp:long_tail_baselines_with_our_gen}
\begin{table*}[t]
  \caption{\textbf{Training long-tailed baselines with generated data from our pipeline.} Logit Adj.~\cite{menon2021logit} and LiVT~\cite{xu2023livt} are trained on K100-LT with and without our generated data ($B=990$).}
  \label{tab:long_tail_baselines_with_our_gen}
  \centering
  \scriptsize
  \setlength{\tabcolsep}{4pt}
  \begin{tabular}{@{} l c 
    >{\columncolor{lt_few}}c >{\columncolor{lt_tail}}c >{\columncolor{lt_head}}c c @{}}
    \toprule
    Method & w/ Our Gen. Data & Few & Tail & Head & Avg C/A \\
    \midrule
    Logit Adj.~\cite{menon2021logit} & \xmark & 55.5 & 68.4 & 78.5 & 65.6 \\
    Logit Adj.~\cite{menon2021logit} & \cmark & \textbf{67.0} & 67.4 & 64.4 & 67.0 \\
    \midrule
    LiVT~\cite{xu2023livt} & \xmark & 47.1 & 68.7 & \textbf{91.0} & 64.7 \\
    LiVT~\cite{xu2023livt} & \cmark & 61.8 & \underline{70.3} & 87.3 & \underline{69.6} \\
    \midrule
    \textbf{Gen2Balance} & \cmark & \underline{62.2} & \textbf{75.0} & \underline{88.4} & \textbf{72.6} \\
    \bottomrule
  \end{tabular}
\end{table*}

We test whether existing long-tail baselines also benefit from our generated data: in Table~\ref{tab:long_tail_baselines_with_our_gen}, we retrain Logit Adj.\ and LiVT on it and evaluate on K100-LT (B{=}990).
Both improve in average accuracy, confirming our generated data is broadly useful, yet both trail Gen2Balance, showing that our training recipe makes better use of the same data. Notably, Logit Adj. gains on average but collapses on head classes ($-14.1\%$), whereas Gen2Balance preserves head accuracy while improving the tail.

\subsection{Evaluating Test-Set Memorisation in WAN}
\label{supp:wan_test_leakage}
Details of the training data for WAN 2.1~\cite{wan2025wan} are not public, so it may have seen K100-LT or UCF-LT test clips and reproduced them in our generated samples. We compute ViCLIP~\cite{wang2024internvid} (video-aligned CLIP) cosine similarity between each generated video and its nearest test video for K100-LT. Generated videos are less similar to the test set than the real training videos themselves: class-averaged $0.677\pm0.093$ vs.\ $0.764\pm0.075$ (cross-class control $0.633$). Since the real training clips share no source with the test set, by design of the Kinetics dataset, the generated videos are further removed from it, showing no leakage.

\section{Dataset Details}
\label{sec:dataset_details}

\subsection{Creation of Long-Tail Action Recognition Datasets}
\label{supp:long-tail-creation}
In Section~\ref{sec:experiments}, we present the statistics of K100-LT and UCF-LT.
Here, we provide additional details on the construction of these long-tailed variants of the Kinetics and UCF-101 datasets.

Following prior long-tail dataset procedures~\cite{liu2019large,perrett2023use,lin2024vlg}, we sample class sizes from a Pareto distribution with $\alpha=5.0$ for K100-LT and $\alpha=6.0$ for UCF-LT, setting the minimum class size to 5. We preserve the original class-size ordering so that the largest class retains its original size.
We then sample training videos from the original datasets randomly up to the new class size.

For K100-LT, we select 100 classes from Kinetics-400, prioritising temporally challenging actions as identified in~\cite{sevilla2021only}. The selected classes, listed in decreasing order of their frequency in K100-LT, are:
{\small \textit{canoeing or kayaking, hammer throw, punching bag, gymnastics tumbling, cheerleading, rock climbing, skiing (not slalom or crosscountry), playing trombone, playing violin, playing tennis, pull ups, bench pressing, throwing discus, capoeira, bowling, swimming backstroke, driving car, belly dancing, ski jumping, smoking, country line dancing, pumping fist, side kick, somersaulting, pole vault, milking cow, roller skating, breakdancing, tap dancing, shaving head, snowboarding, playing accordion, dribbling basketball, playing ice hockey, clean and jerk, playing drums, robot dancing, cleaning floor, opening present, busking, catching or throwing softball, tying knot (not on a tie), kicking field goal, stretching leg, high kick, shuffling cards, kitesurfing, playing didgeridoo, sled dog racing, parasailing, catching or throwing baseball, cutting watermelon, weaving basket, playing cards, writing, drop kicking, playing keyboard, changing oil, cleaning shoes, bouncing on trampoline, swimming butterfly stroke, folding clothes, jumpstyle dancing, krumping, playing cymbals, grooming horse, getting a haircut, throwing ball, hurdling, cartwheeling, shining shoes, mopping floor, drinking, sanding floor, arranging flowers, vault, hoverboarding, planting trees, skiing slalom, ironing, clay pottery making, wrestling, egg hunting, parkour, auctioning, skiing crosscountry, swinging on something, skipping rope, hockey stop, garbage collecting, doing aerobics, changing wheel, building cabinet, gargling, making a sandwich, water sliding, recording music, making tea, swinging legs, drinking shots.}}

\subsection{RareAct Classes Selection}
\label{supp:rareact}
In Table~\ref{tab:rareact} we evaluated our method's ability to recognise 22 rare action classes from the RareAct~\cite{miech2020rareact} dataset. Here we provide additional details of the class selection and training samples. 

RareAct contains 122 rare action classes defined as verb–noun pairs (\eg \textit{drill phone}, \textit{microwave shoes}). 
For each class, a set of clips is annotated from public YouTube videos, often multiple clips from the same video.
Studying the publicly available annotations, we found substantial annotation noise: most clips were unrelated to their assigned class. To clean this dataset, we manually inspect the video clips for noise and discard irrelevant clips (\eg, clips missing the action annotated in the clip). 
After manual cleaning, we retain only classes with at least 50 samples, allowing 5 training examples (appended as few-shot classes to K100-LT) and 45 test samples in the test set. 
We ensure that there is no source-video overlap between the train and test splits when partitioning the dataset. 
This yields 22 classes: \textit{wash chair, cut car, measure pumpkin, cut pumpkin, weigh pumpkin, peel corn, cut keyboard, wash rock, spray door, drill phone, wash pepper, spray fridge, hammer rock, spray pumpkin, hammer car, weigh shoes, hammer phone, spray shoes, weigh tomato, wash window, measure hair}, and \textit{wash potato}. 

As this curated set is too small to serve as a standalone long-tailed benchmark, we append these classes to K100-LT. We release, on our webpage, the curated train/test splits for this clean subset of RareAct.

\begin{figure*}[t]
  \centering
 \includegraphics[width=\linewidth]{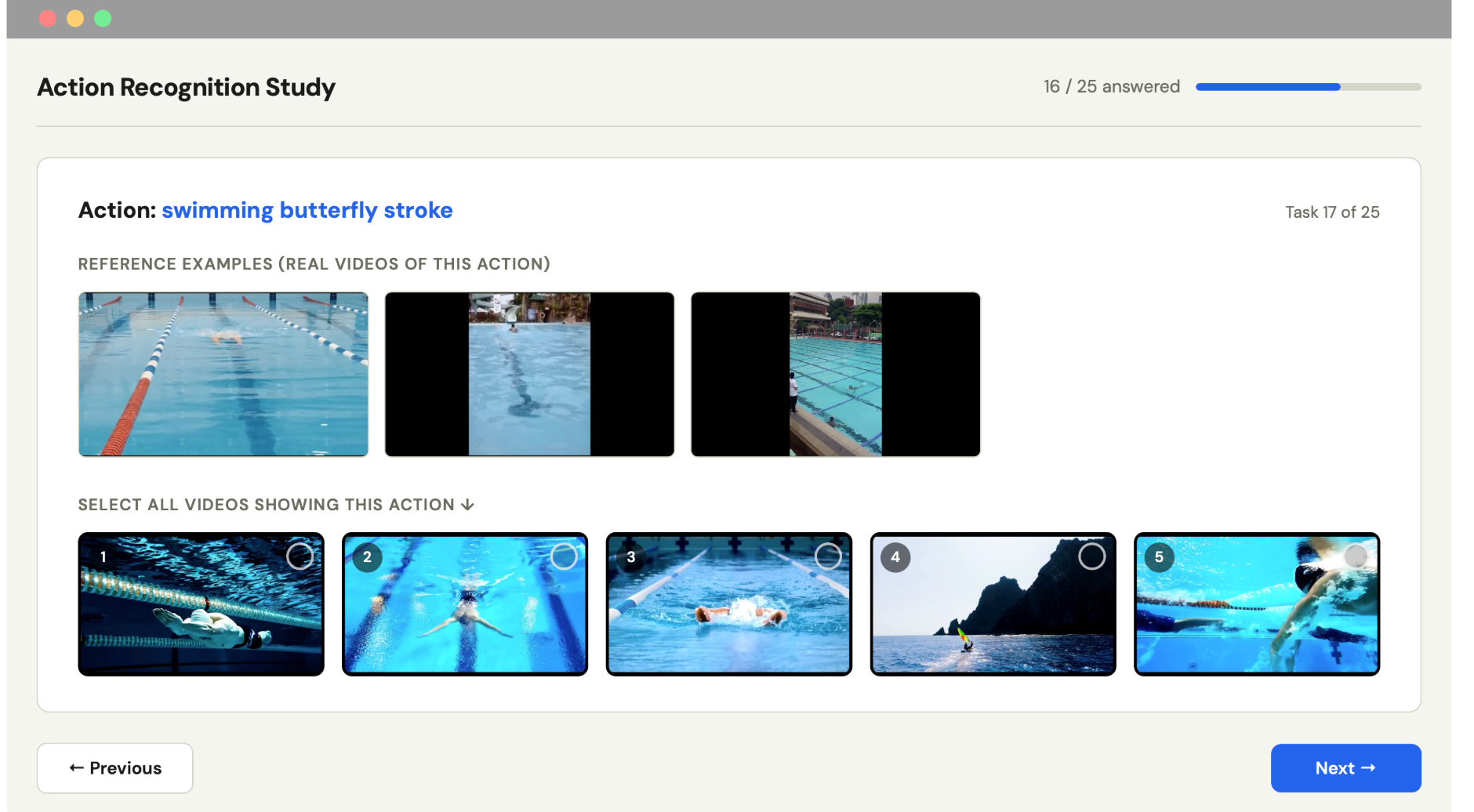}
  \caption{\textbf{User Study Interface}. Users are provided with a target action class and 3 real reference videos to establish semantic grounding. They are then asked to select all candidate-generated videos (out of 5) that accurately depict the target action, rejecting the distractors, testing human recognisability of generated videos. In this example, the fourth video depicts \textit{parasailing} rather than  \textit{swimming butterfly stroke}.}
  \vspace{-13pt}\label{fig:user_study_interface}
\end{figure*}

\begin{figure*}[h!]
  \centering
 \includegraphics[width=\linewidth]{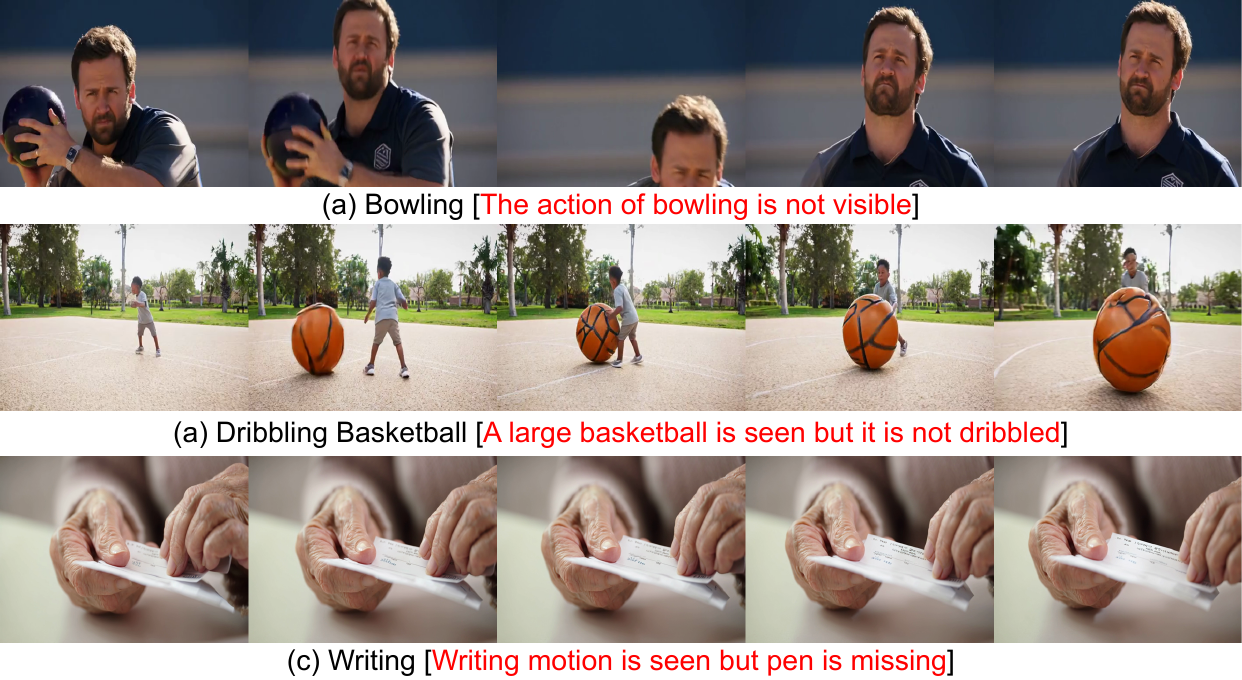}
  \caption{\textbf{Failures in Gen2Balance generated videos.} Common generative errors include rendering the relevant object without the corresponding action (a, b) or simulating the action's motion without the necessary tool (c).}
  \label{fig:gen_user_study_fail}
  \vspace{-3pt}
\end{figure*}

\begin{figure*}[t]
  \centering
 \includegraphics[width=\linewidth]{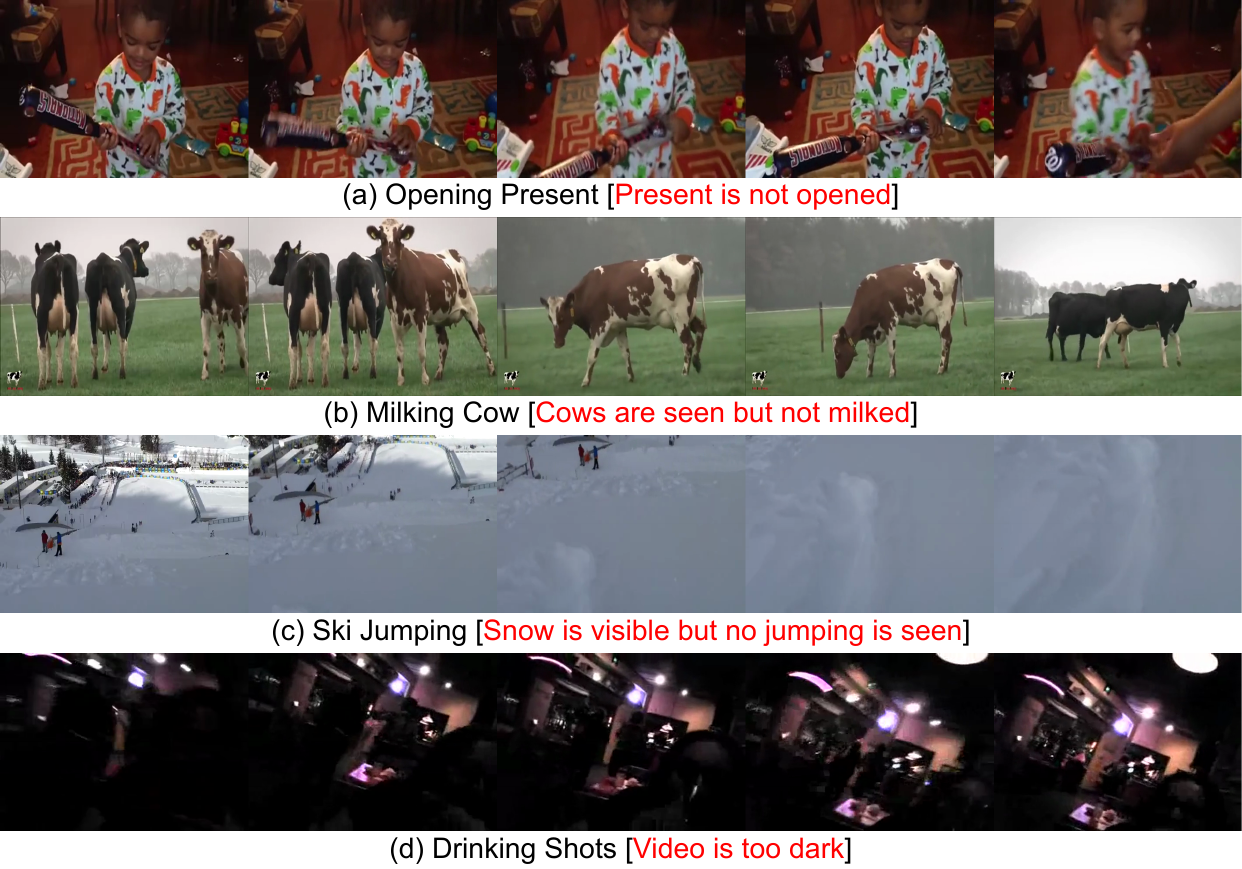}
  \caption{\textbf{Label Noise in real Kinetics-100 videos.} Labelled (ground-truth) noise in these video, detected using our user study, is due to missing actions despite relevant objects or scenes being present (a-c) or severe visibility issues (d).}\vspace{-13pt}\label{fig:real_user_study_fail}
\end{figure*}

\section{User Study of Generated Videos}
\label{supp:user-study}
To assess the semantic quality of videos synthesised by our pipeline, we conduct a user study that mirrors the Kinetics~\cite{kay2017kinetics} curation process, in which annotators judge whether a video depicts a given action class.

In short, we want to evaluate whether the synthesised videos are valid training samples of the labelled class.

\noindent \textbf{User Evaluation of Generated Videos}. We randomly sample $1 \le b \le 5$ video clips from the augmented training data of class $y$. We then randomly sample $c = 5-b$ videos from the generated videos of other classes $\hat{y}\neq{y}$.
This gives us 5 samples for the user to inspect, at least one of which belongs to class $y$.
The number of samples belonging to the class is unknown to the user at each annotation and varies across tasks (each user completed 25 or more annotation tasks).
The user is then asked to select all samples that are representative of the class name $y$.
To avoid misunderstanding, we also show 3 truly labelled examples for that class.
A sample of our interface is shown in Fig.~\ref{fig:user_study_interface}.
We annotate 500 tasks (2,500 individual judgements) by 8 users.

From this user evaluation, the accuracy of our generated videos (\ie, valid class-matchings) was measured at \textbf{87.0\%}. 
All errors here are false negatives - i.e., the user believed the generated video is not a true representation of the class.
No false positives were detected - i.e., a video of a different class being incorrectly selected.
Since the user is making a binary decision about whether a video belongs to the class, the random baseline here is 50\%.

We show a sample of failures - \ie, generated videos deemed not representative of the class in Fig.~\ref{fig:gen_user_study_fail}. These generated videos typically exhibit out-of-view actions (a), object presence without the correct action (b), or simulated/mimicked motion without the necessary tool (c).

\noindent \textbf{User Evaluation of Real Videos}. We conduct an analogous experiment, but for the real videos from Kinetics-100 classes.
For real training videos, the user study resulted in the accuracy of \textbf{92\%}, where all errors were also false negatives.
As shown in Fig.~\ref{fig:real_user_study_fail}, this label noise in Kinetics typically corresponds to the right objects but a missing action (a-c), for example, a wrapped present, but it is not opened in the video, or cows in the field, but are not being milked. Another source of label noise relates to video quality issues, such as extreme darkness (d).

Importantly, compared to this reference user study, our generated videos would be deemed of an acceptable quality with a narrow 5\% gap in noise compared to the label noise in the Kinetics videos.

\section{Gen2Balance LLM Prompts}
\label{sec:supp_prompts}

We provide the full prompts used to query the multimodal LLM $\mathcal{M}$ (Gemini~2.5~Pro) in our generation pipeline (as described in Section~\ref{subsec:gen2bal-generation}).
Section~\ref{sec:supp_prompt_profile} shows the prompt for generating an \emph{Action Profile} $\mathcal{A}_c$ given the class name and in-context video exemplars.
Section~\ref{sec:supp_prompt_diversity} shows the prompt for generating diverse text-to-video prompts $\mathcal{T}_c$ conditioned on the action profile.

\subsection{Action Profile Generation Prompt}
\label{sec:supp_prompt_profile}

The following prompt is sent to $\mathcal{M}$ together with $|S_c|{=}5$ video exemplars from the training set.
The placeholder \texttt{\{action\_class\}} is replaced with the class name at runtime.

\begin{promptbox}[Action Profile Generation Prompt]
\small

\textbf{Prompt for Deconstructing a Visual Action}

\medskip
\textbf{Role:} You are a highly structured expert in visual analysis and semantics. Your purpose is to deconstruct a given human action into its core components for an AI video generation pipeline.

\medskip
\textbf{Goal:} For the given action class: \texttt{\{action\_class\}}, generate a single, structured JSON object that provides a complete visual and semantic definition of the action. In cases where the action class name may have multiple interpretations, use the provided reference videos to determine the specific meaning intended.

\medskip
\textbf{Output Requirements:}
\begin{enumerate}[leftmargin=*, nosep]
    \item \textbf{Format:} The output MUST be a single, raw JSON object.
    \item \textbf{Strictness:} Follow the provided JSON format. Do NOT add any explanatory text, comments, or apologies before or after the JSON object. Your entire response must be the JSON content itself.
    \item \textbf{JSON Schema:} The JSON object must conform to the following schema:
    \begin{itemize}[nosep]
        \item \texttt{"action\_definition": string} -- A concise, one-sentence definition that accurately describes the action using visual cues.
        \item \texttt{"key\_visual\_elements": Array<string>} -- A list of essential objects, body parts, movements, or other visual aspects observed during the action. Be specific.
        \item \texttt{"common\_mistakes\_to\_avoid": Array<string>} -- A list of visually similar but incorrect actions. This requires you to infer what could be confused with the specific action shown.
    \end{itemize}
\end{enumerate}

\medskip
\textbf{Reference Examples:} The following examples demonstrate the expected JSON structure and level of detail.

\medskip
\textbf{Example 1: brushing teeth}
\begin{lstlisting}[style=json]
{
  "action_definition": "The act of moving a small handheld toothbrush
    with bristles back and forth across the teeth inside the mouth,
    to clean the teeth's surfaces.",
  "key_visual_elements": [
    "A toothbrush held in the dominant hand near the mouth.",
    "Toothpaste visible on the toothbrush bristles or around the mouth.",
    "Back-and-forth or circular motions of the toothbrush against teeth.",
    "The person's mouth slightly open to accommodate the toothbrush.",
    "A bathroom sink or mirror typically visible in the background.",
    "Foaming or bubbles around the mouth from toothpaste.",
    "The person spitting into a sink intermittently.",
    "The elbow bent with the forearm raised to bring the toothbrush
     to mouth height."
  ],
  "common_mistakes_to_avoid": [
    "Using a hairbrush or other type of brush on hair.",
    "Eating or drinking something.",
    "Flossing (uses string/floss rather than a brush).",
    "Using mouthwash (involves swishing liquid, not brushing motions).",
    "Brushing or grooming anything other than teeth."
  ]
}
\end{lstlisting}

\textbf{Example 2: playing violin}
\begin{lstlisting}[style=json]
{
  "action_definition": "The act of producing music where a person holds
    a four-stringed wooden stringed instrument against their shoulder and
    chin while drawing a bow across the strings with one hand and pressing
    the strings with fingers of the other hand.",
  "key_visual_elements": [
    "A wooden violin held between the chin and left shoulder",
    "The right arm moving a long wooden bow horizontally across the
     violin strings in smooth, controlled strokes",
    "The left hand positioned on the violin neck with fingers pressing
     down on strings at various positions",
    "The head tilted to the left side to secure the violin against
     the shoulder",
    "The characteristic curved body shape and f-holes of the violin",
    "Precise arm and wrist movements coordinated between both hands"
  ],
  "common_mistakes_to_avoid": [
    "Playing a cello, or double bass (different sizes and playing
     positions)",
    "Playing a guitar or other plucked string instrument",
    "Holding the violin like a guitar against the body",
    "Using fingers to pluck strings instead of bowing",
    "Holding the violin without playing the instrument"
  ]
}
\end{lstlisting}

\textbf{Example 3: playing basketball}
\begin{lstlisting}[style=json]
{
  "action_definition": "The act of participating in a team sport where
    players dribble, pass, and shoot an orange basketball toward an
    elevated basketball hoop while moving around a court, often
    competing against other players.",
  "key_visual_elements": [
    "An orange, textured basketball.",
    "A basketball hoop with a backboard and net.",
    "A person bouncing (dribbling) an orange basketball repeatedly
     against the floor while running or walking.",
    "The act of shooting: propelling the ball towards the hoop with
     a specific form.",
    "Passing the ball between players."
  ],
  "common_mistakes_to_avoid": [
    "Playing with a different type of ball (e.g., a soccer ball,
     football or volleyball).",
    "Simply standing on a basketball court without interacting with
     a ball.",
    "Athletic drills on a court that do not involve a basketball.",
    "Throwing other objects into a hoop.",
    "Players jumping or running on a basketball court without a ball."
  ]
}
\end{lstlisting}

\medskip
\textbf{Your Task:}\\
\textbf{Action Class:} \texttt{\{action\_class\}}\\
\textbf{Reference Content:} Video files are provided with this prompt. Analyse the provided video files to understand the specific visual characteristics of the action class: `\texttt{\{action\_class\}}'. If the action name could have multiple interpretations (e.g., `batting' could mean both baseball or cricket), use the videos to determine the intended meaning.

\end{promptbox}

\subsection{Diverse Text-to-Video Prompt Generation}
\label{sec:supp_prompt_diversity}

The following prompt is sent to $\mathcal{M}$ to generate the set of diverse text prompts $\mathcal{T}_c$ for each class.
The placeholders \texttt{\{action\_class\}} and the action profile fields are filled programmatically. The full set of prompts $\mathcal{T}_c$ is generated in batches of size 25.

\begin{promptbox}[Diverse Prompt Generation Prompt]
\small

\textbf{Instruction for Generating Diverse Text-to-Video Prompts}

\medskip
\textbf{Role:} You are an expert at creating detailed, visually-specific prompts for text-to-video generation models. You understand the importance of precise visual descriptions and action clarity.

\medskip
\textbf{Goal:} Generate 25 diverse, specific text prompts that could be used for a text-to-video generation model to create videos of the action ``\texttt{\{action\_class\}}''.

\medskip
\textbf{Action Profile}\\
Based on the provided action analysis for ``\texttt{\{action\_class\}}'':

\textbf{Definition:} \texttt{\{action\_definition\}}

\textbf{Positive Constraints (do depict):} \texttt{\{key\_visual\_elements\}} {\small(listed as bullet points)}

\textbf{Negative Constraints (do NOT depict):} \texttt{\{common\_mistakes\_to\_avoid\}} {\small(listed as bullet points)}

\medskip
\textbf{Requirements:} 
Each prompt should:
\begin{itemize}[leftmargin=*, nosep]
    \item Explicitly mention the action class (e.g., ``person is eating'' if the action is ``eating'')
    \item Include a brief description of the physical movements involved in the action
    \item Describe the action in a specific context, with specific details
    \item Incorporate the key visual elements identified above naturally into the prompts
    \item Ensure the action depicted is clearly \texttt{\{action\_class\}} and not any of the similar actions to avoid
\end{itemize}

Create a mix of scenarios that naturally fit the action -- focus on the typical contexts where this action actually occurs (e.g., for eating: restaurants, homes, picnics; for exercising: gyms, parks, home workouts; for professional activities: offices, workshops, studios).

\medskip
Vary the following aspects across the set of prompts while maintaining realistic, natural scenarios:
\begin{itemize}[leftmargin=*, nosep]
    \item \textbf{Settings/environments} (realistic locations where the action naturally occurs such as homes, public spaces, outdoors, workplaces, etc.)
    \item \textbf{Camera angles and framing} (close-up, wide shot, side view, shaky handheld, stable tripod, etc.)
    \item \textbf{Video quality} (professional-looking, amateur phone camera, slightly blurry, well-lit, poorly lit, etc.)
    \item \textbf{People performing the action} (different ages, demographics, clothing, single/multiple people)
    \item \textbf{Objects involved} (different tools, items, variations relevant to the action)
    \item \textbf{Action styles} (different speeds, intensities, skill levels from amateur to professional)
    \item \textbf{Time of day and lighting conditions} (daylight, evening, indoors with artificial lighting, etc.)
    \item \textbf{Background elements} (crowded, empty, urban, rural, etc.)
    \item \textbf{Social context} (alone, with family, with friends, in public, etc.)
\end{itemize}

\medskip
\textbf{Format:}
Return exactly 25 prompts as a numbered JSON dictionary with keys ``1'' through ``25''. Each prompt should be a single sentence that is clear, specific, and detailed enough to guide a text-to-video model, typically 15--30 words. Always begin with a subject followed by the action performed (for example, ``A person is eating'') or the appropriate form of the action verb, and include a description of the physical movements.

\medskip
\textbf{Examples:}

\medskip
\textbf{Action Class: drinking}
\begin{lstlisting}[style=json]
{
  "1": "A teenager is drinking water, tilting his head back as he
    raises a plastic bottle to his lips, while skateboarding in
    a park.",
  "2": "An elderly man is drinking hot tea, carefully lifting a
    chipped mug to his mouth with both hands, in his dimly lit
    kitchen with a cluttered counter.",
  "3": "A mother is drinking coffee, taking quick sips between
    conversations, from a travel mug at a playground while
    keeping an eye on her children, viewed from across the
    playground.",
  "4": "Office workers are drinking water, passing a bottle around
    and tipping it back to swallow, during an informal meeting
    in a fluorescent-lit break room.",
  "5": "A sweaty runner is drinking from a sports bottle, squeezing
    liquid into his mouth without touching the rim, after a race
    with cheering spectators in the background, captured in
    close-up.",
  (...)
}
\end{lstlisting}

\textbf{Action Class: kicking field goal}
\begin{lstlisting}[style=json]
{
  "1": "A teenage girl is kicking a field goal, taking three steps
    back before running forward and sending the ball between
    posts with her foot, on an empty school field with makeshift
    posts.",
  "2": "A father is teaching his young son how to kick a field goal,
    demonstrating the proper foot placement and swinging motion,
    in their backyard with the house visible behind them, seen
    from a side angle.",
  "3": "An amateur player is kicking a field goal, planting his
    non-kicking foot firmly while swinging his kicking leg
    through the ball, at a community game with a small crowd
    watching.",
  "4": "A professional football player is kicking a field goal,
    executing a perfect follow-through as his foot connects with
    the ball sending it high above the crossbar, during a
    televised game viewed from behind the goalposts.",
  "5": "Friends are taking turns kicking field goals, setting up
    the ball on a backpack and attempting to kick it between two
    trees, in a park on a sunny afternoon."
  (...)
}
\end{lstlisting}

\textbf{Action Class: playing guitar}
\begin{lstlisting}[style=json]
{
  "1": "A young woman is playing guitar, her fingers pressing down
    on frets while her other hand strums the strings, alone in
    her bedroom with basic lighting.",
  "2": "A street performer is playing an acoustic guitar, rhythmically
    tapping the body while picking individual strings, in a busy
    pedestrian area with people walking by, shown from across
    the street.",
  "3": "A grandfather is playing guitar, slowly demonstrating chord
    changes by positioning his fingers on the fretboard, while
    teaching his grandchild on an old instrument.",
  "4": "A teenager is playing electric guitar, head bobbing as he
    rapidly moves his picking hand up and down across the
    strings, in a garage band practice, filmed from a low
    angle.",
  "5": "Fingers are playing intricate riffs on a classical guitar,
    precisely moving between strings and frets with controlled
    movements, in a close-up view."
  (...)
}
\end{lstlisting}

\medskip
\textbf{Important Guidelines Based on Action Analysis}\\
When generating prompts for ``\texttt{\{action\_class\}}'':
\begin{enumerate}[leftmargin=*, nosep]
    \item Ensure each prompt clearly depicts the specific action as defined, not similar activities
    \item Include relevant visual elements naturally in the scene descriptions
    \item Be specific about the movements and objects involved to avoid ambiguity
    \item Double-check that your prompts won't be confused with the similar actions listed above
\end{enumerate}

\medskip
Please generate exactly 25 unique, diverse prompts for the action ``\texttt{\{action\_class\}}'' following these patterns. Make sure to return exactly 25 prompts numbered from 1 to 25.
\end{promptbox}

\section{Sample Action Profiles}
\label{supp:action-profiles}

We show sample action profiles produced by our pipeline (Sec.~\ref{subsec:gen2bal-generation}), one per benchmark: Robot Dancing (K100-LT), Handstand Walking (UCF-LT), and Cut Keyboard (RareAct). Each profile was generated by prompting Gemini 2.5 Pro with the class name and five real video exemplars from the training set (corresponding to the final stage in Fig.~\ref{fig:method_generation_pipeline}). A profile comprises a definition, positive constraints (key visual elements to include), and negative constraints (common mistakes to avoid). These profiles then condition the diverse prompt generation stage to produce class-faithful videos.

\begin{actionprofile}{Action Profile for Robot Dancing {\small\textit{(K100-LT)}}}
\textbf{Action Definition.} A person dancing characterised by stiff, jerky, and isolated movements of the limbs and body, intended to imitate the mechanical motions of a robot or automaton.

\medskip
\textbf{Positive Constraints ({\color{green!40!black}Key Visual Elements}).}
\begin{itemize}[nosep, leftmargin=1.2em]
    \item Stiff and rigid body posture.
    \item Jerky, staccato movements with clear start and stop points.
    \item Isolation of body parts where one part moves while others remain still.
    \item Locking and popping of joints (elbows, wrists, knees).
    \item Arms bent at sharp geometric angles.
    \item Slow, deliberate, and controlled motions.
    \item A lack of fluid transitions.
\end{itemize}

\medskip
\textbf{Negative Constraints ({\color{red!60!black}Mistakes to Avoid}).}
\begin{itemize}[nosep, leftmargin=1.2em]
    \item Fluid dance styles (e.g., contemporary, ballet).
    \item Breakdancing with floor work and spins.
    \item Waving or tutting.
    \item Simple non-rhythmic gesturing.
    \item Random flailing lacking deliberate control.
\end{itemize}
\end{actionprofile}

\begin{actionprofile}{Action Profile for Handstand Walking ({\small\textit{UCF-LT}})}
\textbf{Action Definition.} The act of supporting the entire body weight on the hands in an inverted, vertical position and moving forward by alternately lifting and placing each hand on the ground in a walking motion.

\medskip
\textbf{Positive Constraints ({\color{green!40!black}Key Visual Elements}).}
\begin{itemize}[nosep, leftmargin=1.2em]
    \item A person's body is inverted, with feet and legs above the head and torso.
    \item Entire body weight supported by arms and hands on the ground.
    \item Forward locomotion through alternating ``stepping'' motion of the hands.
    \item Arms kept mostly straight to support the body's weight.
    \item Legs held in the air---straight, bent, or split to maintain balance.
    \item Head positioned close to the ground, between the arms.
\end{itemize}

\medskip
\textbf{Negative Constraints ({\color{red!60!black}Mistakes to Avoid}).}
\begin{itemize}[nosep, leftmargin=1.2em]
    \item A static handstand without forward movement.
    \item Cartwheels or roundoffs.
    \item Handstand push-ups (vertical motion only, no forward travel).
    \item Falling or stumbling immediately after entering a handstand.
    \item Crawling on all fours with the body facing the ground.
\end{itemize}
\end{actionprofile}

\begin{actionprofile}{Action Profile for Cut Keyboard {\small\textit{(RareAct)}}}
\textbf{Action Definition.} The act of using a sharp tool, often a power tool like an angle grinder, to forcefully sever the plastic casing and components of a computer keyboard.

\medskip
\textbf{Positive Constraints ({\color{green!40!black}Key Visual Elements}).}
\begin{itemize}[nosep, leftmargin=1.2em]
    \item A standard computer keyboard being held stationary, often by a clamp.
    \item A cutting tool, most prominently an angle grinder with a spinning disc.
    \item The blade making contact with and moving through the keyboard's plastic casing and keys.
    \item Sparks, smoke, and plastic debris generated at the point of contact.
    \item A person wearing protective gear (e.g., gloves) while operating the tool.
    \item The keyboard being split into two or more pieces as a result of the cut.
\end{itemize}

\medskip
\textbf{Negative Constraints ({\color{red!60!black}Mistakes to Avoid}).}
\begin{itemize}[nosep, leftmargin=1.2em]
    \item Typing on a keyboard.
    \item Cleaning a keyboard with brushes or compressed air.
    \item Disassembling a keyboard non-destructively with a screwdriver.
    \item Cutting a different object like wood or metal.
    \item Prying open the keyboard case with a dull object.
    \item Smashing a keyboard with a hammer.
\end{itemize}
\end{actionprofile}

\section{Sample Text Prompts}
\label{supp:text-prompts}

We show sample text prompts generated by conditioning on the action profiles (Sec.~\ref{supp:action-profiles}) and diversity axes described in Sec.~\ref{subsec:gen2bal-generation}. For each class, we highlight five prompts that illustrate variation across different axes.

\begin{actionprofile}{Text prompts for Robot Dancing (K100-LT)}
\begin{enumerate}[nosep, leftmargin=1.5em, label=\textbf{\arabic*.}]
    \item ``A child in a silver costume is robot dancing on a school 
          stage, performing simple, staccato arm movements and rigid 
          poses for an audience.''
    \item ``A street performer is robot dancing for a small crowd, 
          executing sharp, isolated pops with his arms and chest in 
          perfect time to a boombox beat.''
    \item ``Filmed on grainy VHS, a teenager is robot dancing in 
          their bedroom, showcasing stiff, isolated arm movements 
          and abrupt locking poses to 90s electronic music.''
    \item ``From a low angle, a dancer is robot dancing on a 
          basketball court, their powerful, jerky movements and 
          body pops looking dramatic against the chain-link fence.''
    \item ``A man in an LED suit is robot dancing on a dark street 
          at night, his stiff, isolated movements and abrupt stops 
          visually traced by the glowing lights on his costume.''
\end{enumerate}
\end{actionprofile}

\begin{actionprofile}{Text Prompts for Handstand Walking (UCF-LT)}
\begin{enumerate}[nosep, leftmargin=1.5em, label=\textbf{\arabic*.}]
    \item ``A muscular man is handstand walking across a rubber gym 
          floor, propelling himself forward by rapidly alternating 
          hand placements during a CrossFit workout.''
    \item ``A young child is attempting to handstand walk in a 
          carpeted living room, taking a few wobbly forward steps 
          with his hands before his inverted body tips over.''
    \item ``From a low-angle tracking shot, a fit person is handstand 
          walking along the wet sand at sunset, advancing by 
          rhythmically planting one hand after the other.''
    \item ``Grainy, vintage-style footage shows a circus strongman 
          handstand walking up a slight incline, his body supported 
          on his hands as he makes deliberate forward progress.''
    \item ``A drone shot follows a person handstand walking along a 
          scenic cliffside path, their small, inverted figure moving 
          forward against a vast ocean backdrop.''
\end{enumerate}
\end{actionprofile}

\begin{actionprofile}{Text Prompts for Cut Keyboard (RareAct)}
\begin{enumerate}[nosep, leftmargin=1.5em, label=\textbf{\arabic*.}]
    \item ``A person is cutting a keyboard with a fast, aggressive 
          motion, slicing through it in one quick pass that sends 
          large chunks of plastic keys flying across the floor.''
    \item ``A hobbyist is cutting a keyboard for a custom mod, using 
          a fine-toothed hand saw to carefully remove the number pad 
          section on a desk covered in tools and wires.''
    \item ``Filmed in slow motion, a person is cutting a keyboard, 
          and tiny fragments of plastic keys and shavings explode 
          outwards as a saw blade bites into the device.''
    \item ``In a dimly lit garage, a man is cutting a keyboard 
          clamped to a workbench, the bright flash of sparks from 
          the tool briefly illuminating the dusty room.''
    \item ``From a top-down perspective, an artist is cutting a 
          keyboard, carefully slicing between the key rows with a 
          craft knife to separate pieces for a sculpture.''
\end{enumerate}
\end{actionprofile}

\bibliographystylesuppl{splncs04}

\end{document}